\documentclass[sort]{commtr}
\usepackage{caption}
\usepackage{makecell}
\usepackage{multirow}
\usepackage{enumitem}
\usepackage{ulem}
\usepackage{subfigure}
\usepackage{url}

\usepackage{booktabs}
\usepackage{multirow}
\usepackage{xspace}
\usepackage{amsmath}
\usepackage{color}
\usepackage{threeparttable}
\usepackage[ruled,linesnumbered]{algorithm2e}
\usepackage{algorithm2e}
\usepackage{amsmath} 
\usepackage{amsfonts} 
\usepackage{threeparttable}

\usepackage[utf8]{inputenc}
\usepackage[T1]{fontenc}
\usepackage{booktabs}

\newtheorem{definition}{Definition}

\newcommand{\eat}[1]{}

\let\oldhat\hat

\renewcommand{\hat}[1]{\oldhat{\mathbf{#1}}}

\commtrsetup{
title    = {HUTFormer: Hierarchical u-net transformer for long-term traffic forecasting},
author   = {Zezhi Shao$^{\rm a}$, Fei Wang$^{\rm a,b,*}$, Tao Sun$^{\rm a}$, Chengqing Yu$^{\rm a,b}$, Yuchen Fang$^{\rm c}$, Guangyin Jin$^{\rm d}$, Zhulin An$^{\rm a}$, Yang Liu$^{\rm e}$, Xiaobo Qu$^{\rm e}$, Yongjun Xu$^{\rm a,b}$},
address  = {
\tdd{a} {Institute of Computing Technology, Chinese Academy of Sciences, Beijing, 100190, China} \sep
\tdd{b} {School of Computer Science and Technology, University of Chinese Academy of Sciences, Beijing, 100049, China} \sep
\tdd{c} {School of Computer Science and Engineering, University of Electronic Science and Technology of China, Chengdu, 610054, China} \sep
\tdd{d} {Department of Planning, Design, and Technology of Architecture, Sapienza University of Rome, Rome, 00196, Italy} \sep
\tdd{e} {School of Vehicle and Mobility, Tsinghua University, Beijing, 100084, China} \sep
\tdd{*} Corresponding author.
},
email    = {wangfei@ict.ac.cn},
abstract = {
Traffic forecasting, which aims to predict traffic conditions based on historical observations, has been an enduring research topic and is widely recognized as an essential component of intelligent transportation.
Recent proposals on Spatial-Temporal Graph Neural Networks~(STGNNs) have made significant progress by combining sequential models with graph convolution networks. However, due to high complexity issues, STGNNs only focus on short-term traffic forecasting (e.g., 1-h ahead), while ignoring more practical long-term forecasting. In this paper, we make the first attempt to explore long-term traffic forecasting (e.g., 1-day ahead). To this end, we first reveal its unique challenges in exploiting multi-scale representations. Then, we propose a novel Hierarchical U-Net TransFormer~(HUTFormer) to address the issues of long-term traffic forecasting. HUTFormer consists of a hierarchical encoder and decoder to jointly generate and utilize multi-scale representations of traffic data. Specifically, for the encoder, we {\color{black}propose} window self-attention and segment merging to extract multi-scale representations from long-term traffic data. For the decoder, we design a cross-scale attention mechanism to effectively incorporate multi-scale representations.  In addition, HUTFormer employs an efficient input embedding strategy to address the complexity issues. Extensive experiments on four traffic datasets show that the proposed HUTFormer significantly outperforms state-of-the-art traffic forecasting and long time series forecasting baselines.
},
keywords = {
traffic condition forecasting, long-term time series forecasting, multivariate time series forecasting
},
}

\begin{document}

\maketitle  


\Nomenclature

\begin{center}
\begin{tabular}{|P{.03\linewidth}|P{.85\linewidth}|}
\hline
      $T$ & The length of history traffic data \\
      $T_\mathrm{f}$ & The length of future traffic data\\
      $N$ & The number of time series \\
      $C$ & The number of feature channels in a traffic sensor\\
      $\mathcal{X}$ & History data of shape $\mathbb{R}^{T\times N\times C}$\\
      $\mathcal{Y}$ & Future data of shape $\mathbb{R}^{T_f\times N\times C}$\\
      $\mathbf{X}^i$ & History data of sensor $i$\\
      $\mathbf{Y}^i$ & Future data of sensor $i$\\
      $\hat{\mathbf{Y}}^i$ & Prediction data of sensor $i$\\
      $L$ & The segment size\\
      $P$ & The number of segments. $T=P\times L$\\
      $d$ & The hidden dimension\\
      $\mathbf{W}$ & Parameter matrix of the fully connected layer\\
      $\mathbf{b}$ & Parameter of the bias of the fully connected layer\\
      $\mathbf{E}$ & Spatial embeddings of shape $\mathbb{R}^{N\times d_1}$\\
      $\mathbf{T}$ & Temporal embeddings\\
      $N_D$ & The number of time slots of a day\\
      $N_W$ & The number of days in a week\\
      $\mathbf{S}$ & Embeddings of each segment after segment embedding\\
      $\mathbf{U}$ & Embeddings of each segment after spatial-temporal positional encoding\\
      $\mathbf{H}$ & Hidden states\\
\hline
\end{tabular}
\end{center}

\section{Introduction}

\label{section:intro}

Traffic forecasting aims at predicting future traffic conditions (e.g., traffic speed or flow) based on historical traffic conditions observed by sensors.
With the development of Intelligent Transportation Systems~(ITS), traffic forecasting fuels a wide range of services related to traffic scheduling, public safety, etc.~\citep{COMMTR1,COMMTR2, TRE1, Innovation4, IF}.
For example, predicting long-term traffic changes (e.g., 1-day) is valuable for people to plan their route in advance to avoid possible traffic congestion.

In general, traffic data\footnote{{\color{black}For the sake of brevity, in this paper, we use `traffic condition data' and `traffic data' interchangeably.}} {\color{black}is presented in the form of multiple time series}, where each time series records traffic conditions observed by sensors deployed on a road network.
A critical property of traffic data is that there exist strong correlations between time series owing to the connection of road networks.
To make accurate traffic forecasting, state-of-the-art proposals~\citep{2018DCRNN,2019GWNet,JGY2} usually adopt Spatial-Temporal Graph Neural Networks (STGNNs), which model the correlations between time series based on Graph Convolution Networks~(GCNs)~\citep{2016GCN,2017GCN,2018DCRNN}.
However, graph convolution brings significant improvements in performance and 50
complexity at the same time. Computational complexity usually increases linearly or quadratically with the length and number of time series~\citep{2022STEP}.
Therefore, it is difficult for STGNNs to scale to long-term historical traffic data, let alone predict long-term future traffic conditions.
In fact, most existing works focus on short-term traffic prediction, e.g., predicting future 12 time steps~(1 h in commonly used datasets).
Such an inability to make long-term traffic forecasting limits the practicality of these models.

\begin{figure}[htp]
  \centering
  \includegraphics[width=1\linewidth]{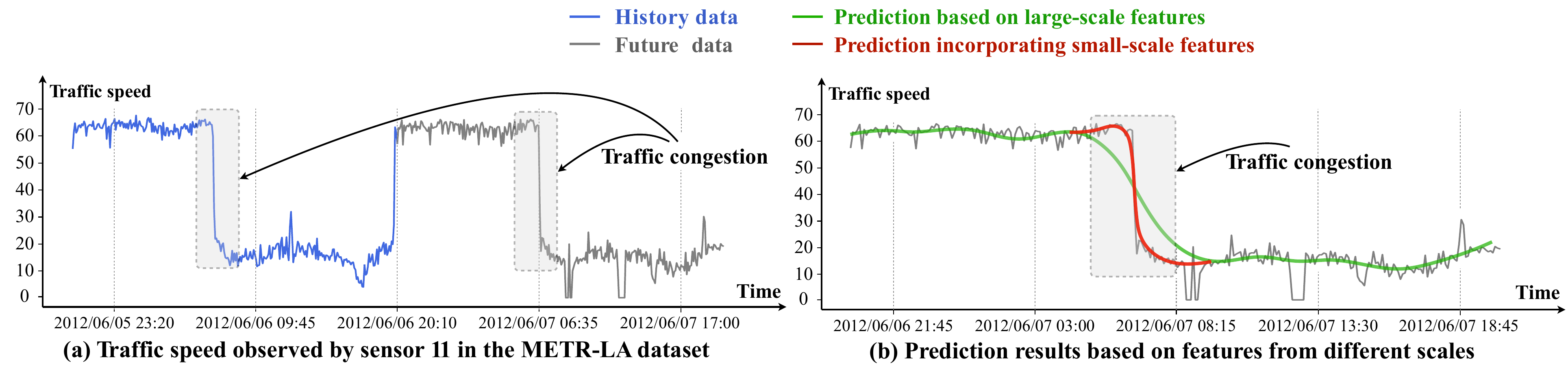}
  \caption{
  Examples of long-term traffic forecasting.}
  \label{figure:intro}
\end{figure}

In this study, we focus on long-term traffic forecasting, e.g., predicting a future day.
{\color{black}
Except for the correlations between time series, the long-term traffic forecasting task has its own uniqueness.}
In the following, we discuss them in detail to motivate model design.
Examples of traffic data is shown in Fig. \ref{figure:intro}\footnote{The Fig. \ref{figure:intro}b is the future part of the Fig. \ref{figure:intro}a.}.
On the one hand, when observing from a global perspective, traffic data usually exhibit regular changes, e.g., daily periodicity.
On the other hand, local details are crucial for traffic forecasting. 
For example, we must capture the rapidly decreasing traffic changes when daily traffic congestion occurs.
{\color{black}
To capture different patterns, exploiting multi-scale representations of traffic data is the key challenge of accurate long-term traffic forecasting.}
Specifically, smaller-scale and larger-scale representations are extracted based on smaller and larger receptive fields, respectively.
The former is usually semantically weak but fine-grained, which facilitates the prediction of local details, e.g., rapid changes during traffic congestion.
In contrast, the latter is coarse-grained but semantically strong, which is helpful in predicting global changes, e.g., daily periodicity. 
An illustration is shown in Fig. \ref{figure:intro}b. 
The prediction based on large-scale features captures daily periodicity but misses local details, which can be fixed by further incorporating small-scale features.

However, it is a challenging task to exploit multi-scale representations of traffic data.
We discuss it from two aspects: Generating and utilizing multi-scale representations.
On the one hand, most existing models cannot generate multi-scale representations of traffic data.
State-of-the-art models for long time series forecasting~\citep{2021Informer} mainly adopt Transformers to capture the long-term dependencies based on self-attention mechanisms~\citep{2017Transformer}.
However, standard self-attention naturally has a global receptive field and thus can only generate representations on a fixed scale.
On the other hand, utilizing multi-scale representations for traffic forecasting is also a challenging task, as it usually requires a specific decoder.
For example, in computer vision tasks like object detection and semantic segmentation, researchers designed decoders such as FPN~\citep{2017FPN} and U-Net~\citep{2015UNet} to utilize the multi-scale representations extracted by the pre-trained encoder~\citep{2016ResNet}.
{\color{black}
These architectures usually require pixel alignment of input and output images.
However, the historical and future sequences in traffic forecasting problems are not the same sequences, i.e., not aligned, making existing approaches~\citep{2017FPN,2015UNet} inapplicable.}

Based on the above discussion, we summarize three challenges that the desired long-term traffic forecasting model should address.
{\color{black}
First, it must efficiently model the correlations between multiple long-term time series.
Second, it should generate multi-scale representations of traffic data by an encoder.
Third, it should include a decoder for traffic forecasting tasks to effectively utilize the multi-scale representations generated by the encoder.}

To address the above challenges, we propose a novel Hierarchica U-net TransFormer (named HUTFormer).
As shown in Fig. \ref{fig:main_arch}, HUTFormer is a two-stage model consisting of a hierarchical encoder and a hierarchical decoder, forming an inverted U-shaped structure.
{\color{black}
To address the efficiency problem}, HUTFormer designs an efficient input embedding strategy, which employs segment embedding and spatial-temporal positional encoding to significantly reduce the complexity of modeling multiple long-term time series in both temporal and spatial dimensions.
{\color{black}
To generate multi-scale representations}, the HUTFormer encoder {\color{black}proposes} a window Transformer layer to limit the receptive field, and then {\color{black}designs} segment merging as a pooling layer to extract larger-scale features.
Thus, lower layers of the encoder focus on {\color{black}smaller-scale} features, while higher layers generate {\color{black}larger-scale} features.
Then, HUTFormer makes an intermediate prediction based on the top-level representations.
{\color{black}
To utilize multi-scale representations, the HUTFormer decoder proposes a cross-scale attention mechanism to address the misalignment issue}, which retrieves information for each segment of the intermediate prediction from multi-scale representations, thus enabling the fine-tuning of the intermediate prediction.
By exploiting the multi-scale representations of traffic data, HUTFormer is capable of making accurate long-term traffic forecasting. 
The main contributions of this paper are summarized as follows:
\begin{itemize}
    \item To our best knowledge, this is the first attempt to study long-term traffic forecasting. 
     We {\color{black}reveal} its unique challenges in exploiting multi-scale representations of traffic data, and propose a novel Hierarchical U-net TransFormer~(HUTFormer) to address them.
    \item 
    We {\color{black}propose} window self-attention and cross-scale attention mechanisms to generate and utilize multi-scale representations effectively. 
    In addition, to address complexity issues, we design an input embedding strategy that includes segment embedding and spatial-temporal positional encoding.
    \item Extensive experiments on four traffic datasets show that the proposed HUTFormer significantly outperforms state-of-the-art traffic forecasting and long-sequence time series forecasting baselines, and effectively exploits the multi-scale representations of traffic data.
\end{itemize}

\section{Related work}
\subsection{Traffic forecasting}
Previous traffic forecasting studies usually fall into two categories, i.e., knowledge-driven~(e.g., queuing theory) and early data-driven models~\citep{kumar2015short, 2014GRU, FC-LSTM, 2016TCN, 2021TITS, 2020AppliedIntelligence,liu2019deeppf,liu2021deeptsp}.
However, these methods usually ignore the correlation between time series and the high non-linearity of time series~\citep{2022D2STGNN}, which severely limits the effectiveness of these methods.
With the development of deep learning~\citep{innovation1,innovation2,innovation3}, Spatial-Temporal Graph Neural Networks (STGNNs) were proposed recently~\citep{2018DCRNN} to model the complex spatial-temporal correlations in traffic data.
Specifically, STGNNs combine Graph Neural Networks~(GNNs)~\citep{2017GCN,2016GCN, HetGNN} and sequential models~(e.g., CNN~\citep{2016TCN} or RNN~\citep{2014GRU}), to model the complex spatial-temporal correlation in traffic data.
For example, DCRNN~\citep{2018DCRNN}, ST-MetaNet~\citep{2019STMetaNet}, AGCRN~\citep{2020AGCRN}, etc.~\citep{2021GTS,2018STGCN,2021DGCRN}, are RNN-based methods, which combine GNN with recurrent neural networks.
Graph WaveNet~\citep{2019GWNet}, MTGNN~\citep{2020MTGNN}, STGCN~\citep{2018STGCN}, and StemGNN~\citep{2020StemGNN} are CNN-based methods~\citep{2021DMSTGCN}, which usually combines GNN with the Temporal Convolution Network (TCN~\citep{2016TCN}).
Moreover, techniques such as attention mechanisms and spectral theories are also widely employed in STGNNs~\citep{2020GMAN, 2020STGNN, 2021ASTGNN,2019ASTGCN,2020STGRET,GinAR, DFGCN,BasicTS}.

Although STGNNs have achieved significant progress, their complexity is high.
Specifically, their complexity usually increases linearly or quadratically with the length and the number of time series~\citep{2022STEP}, since they need to deal with both temporal and spatial dependency at every step.
Thus, most of them focus on short-term traffic forecasting based on short-term history data, e.g., predicting future 1-h traffic conditions based on 1-h historical data~\citep{2018DCRNN, 2019GWNet, 2022D2STGNN, 2022STEP, 2020MTGNN, bogaerts2020graph}.
A recent work STEP~\citep{2022STEP} attempts to address this issue based on the time series pre-training model.
It significantly enhances STGNN's ability to exploit long-term historical data.
However, STEP requires a downstream STGNN as the backbone, which still focuses on short-term traffic forecasting.

Although STGNN-based traffic forecasting has made significant progress, these models only focus on short-term traffic forecasting, and cannot handle long-term traffic forecasting.
On the one hand, due to the high complexity, most of them can not handle long-term history data, let alone predict long-term future traffic conditions.
On the other hand, apart from efficiency issues, long-term traffic forecasting also has its unique challenges, which require exploiting multi-scale representations of traffic data to capture the complex long-term traffic dynamics.



\subsection{Long-sequence time series forecasting}
Recently, long-sequence time series forecasting~\citep{2021Informer} has received much attention~\citep{2021Informer,2021AutoFormer,2022FEDFormer,2022Pyraformer,2023DLinear}.
They aim to make long-term future predictions by modeling long-term historical sequences.
For example, Informer~\citep{2021Informer} proposes a ProbSparse self-attention mechanism to replace the standard self-attention, which enhances the predictive ability of the standard Transformer in the long-sequence forecasting problem.
Autoformer~\citep{2021AutoFormer} designs an efficient Auto-Correlation mechanism to conduct dependencies discovery and information aggregation at the series level.
DLinear~\citep{2023DLinear} rethinks Transformer-based techniques and proposes a simple linear model based on decomposition to achieve better accuracy.
{\color{black}Recently, many advanced Transformer-based models have been proposed, such as PatchTST~\citep{2023PatchTST}, Crossformer~\citep{2023Crossformer}, Scaleformer~\citep{Scaleformer}, and DSformer~\citep{DSformer}.}

Although these models have made considerable progress in long-term time series forecasting, they are not designed for traffic data, which significantly affects their effectiveness in traffic forecasting problems.
We discuss it from two aspects.
First of all, there are strong correlations between multiple time series in traffic data, which is an important bottleneck in traffic forecasting~\citep{2018DCRNN}.
However, long-sequence time series forecasting models usually do not pay attention to such spatial correlations~\citep{2021Informer,2021AutoFormer,2022FEDFormer,2022Pyraformer,Triformer, Scaleformer}{\color{black}, or are not efficient enough~\citep{2023Crossformer}.}
Second, as discussed in Section \ref{section:intro}, long-term traffic forecasting requires exploiting multi-scale representations of traffic data to capture the complex long-term traffic dynamics.
However, long-sequence forecasting models usually rely on the self-attention mechanism and its variants, which can not explicitly generate multi-scale features{\color{black}~\citep{2021Informer,2021AutoFormer,2023PatchTST}}.

\label{sec:ltsf}
\section{Preliminaries}

In this section, we define the notions of traffic data and traffic forecasting task.
\begin{definition}
Traffic data $\mathcal{X}\in\mathbb{R}^{T\times N\times C}$ denotes the observation from all sensors on the traffic network, 
where $T$ is the number of time steps, $N$ is the number of traffic sensors, and $C$ is the number of features collected by sensors.
We additionally denote the data from the sensor $i$ as $\mathbf{X}^{i}\in\mathbb{R}^{T\times C}$.
\end{definition}

\begin{definition}
Traffic forecasting aims to predict the traffic values $\mathcal{Y}\in\mathbb{R}^{T_f\times N\times C}$ of the $T_f$ nearest future time steps based on the given historical traffic data $\mathcal{X}\in\mathbb{R}^{T_h\times N\times C}$ from the past $T_h$ time steps.
In this study, we focus on long-term traffic forecasting, e.g., forecasting for a day in the future.

\end{definition}

\section{Model architecture}

\begin{figure*}[t]
  \centering
  \includegraphics[width=1\linewidth]{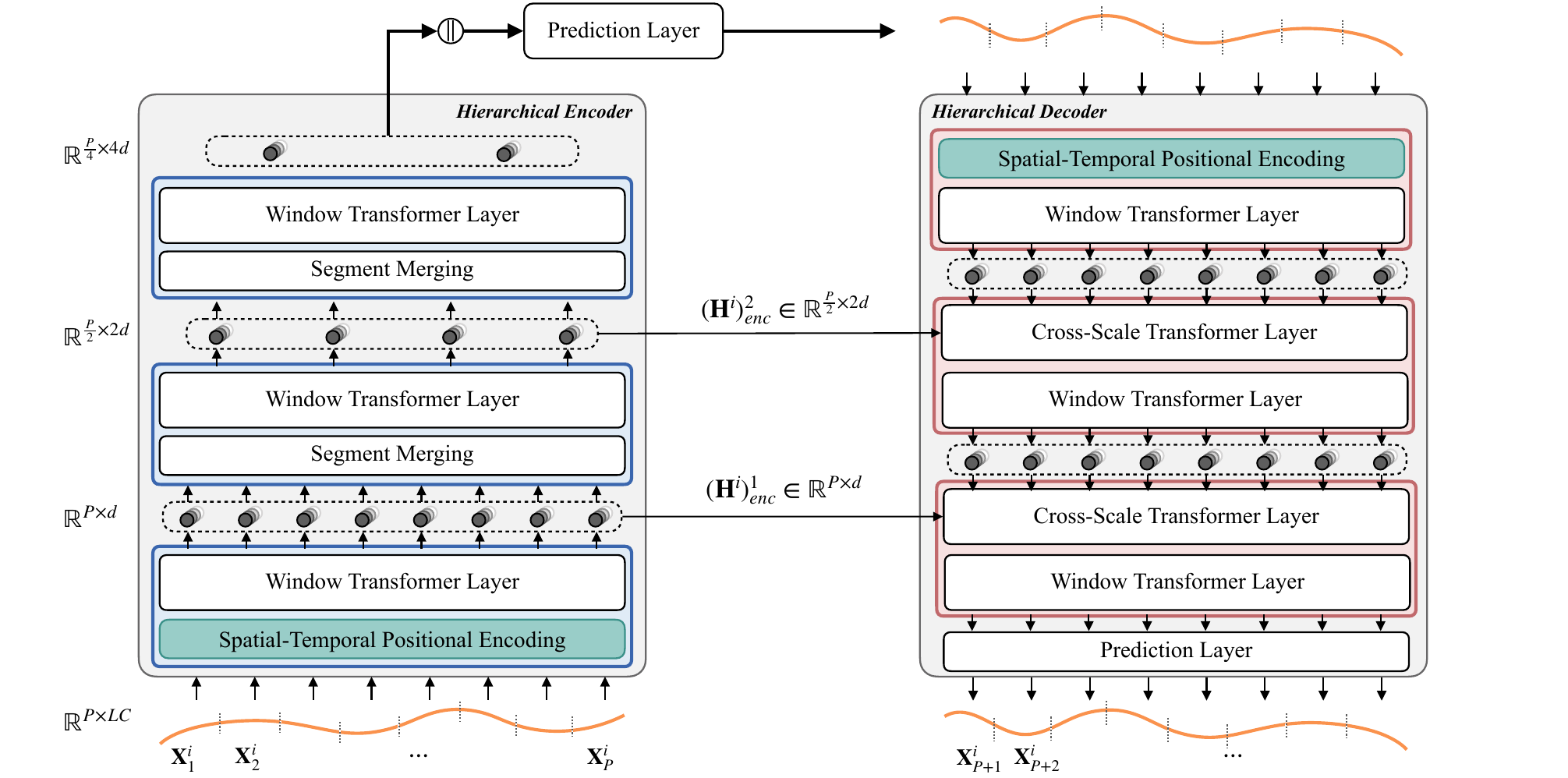}
  \caption{
    Overview of the proposed HUTFormer.
  Left: The hierarchical encoder. It generates multi-scale features for traffic data based on window Transformer layer and segment merging, and makes an intermediate prediction.
  Right: The hierarchical decoder. It fine-tunes the intermediate prediction by incorporating multi-scale features based on cross-scale Transformer layer.
  In addition, segment embedding and spatial-temporal positional encoding are proposed to address complexity issues.
  }
  \label{fig:main_arch}
\end{figure*}

\subsection{Overview}
\label{sec:methods:overview}
As illustrated in Fig. \ref{fig:main_arch}, HUTFormer is based on a hierarchical U-net structure to generate and utilize multi-scale representations of traffic data.
In this subsection, we intuitively discuss each component of HUTFormer and its two-stage training strategy.

First, we discuss the hierarchical encoder.
The window Transformer layer is the basis for generating multi-scale representations, {\color{black}which calculates self-attention within a small window to limit the receptive field.}
Then, segment merging acts as a pooling layer, reducing the sequence length to produce larger-scale representations.
By combining them, lower layers can focus on smaller-scale features while higher layers focus on larger-scale features, thus successfully generating multi-scale features.
Then, an intermediate prediction is made based on the top-layer representations.
However, considering that the top-layer features are semantically strong but coarse-grained, 
the intermediate prediction {\color{black}may fail to capture rapidly changing local details, e.g., the red line in Fig. \ref{figure:intro}b.}

{\color{black}To address the above problem}, the hierarchical decoder aims to fine-tune the intermediate prediction by incorporating multi-scale representations.
U-net~\citep{2015UNet,2021SwinUNet} is a popular structure for utilizing multi-scale representations, especially in computer vision tasks~(e.g., semantic segmentation).
In these tasks, the pixels of the input and target images are aligned, i.e., models operate on the same image.
However, for traffic forecasting tasks, 
the input and output sequences are {\color{black}not the same sequence, i.e.,} not aligned.
Thus, the representations generated by the encoder and the decoder cannot be directly superimposed as regular U-net structures~\citep{2015UNet, 2021SwinUNet} do for computer vision tasks.
To this end, we design a cross-scale Transformer layer, which uses the representations from the decoder as queries and the multi-scale features from the encoder as keys and values to retrieve information.
Such a top-down pathway and lateral connects help to combine the multi--scale representations, thus enhancing the prediction accuracy.

In addition, HUTFormer addresses complexity issues based on an efficient input embedding strategy, which consists of segment embedding and spatial-temporal positional encoding.
On the one hand, segment embedding {\color{black}reduces complexity from the temporal dimension by using} time series segments as basic input tokens.
This simple operation has significant benefits in both reducing the length of the input sequence and providing more robust semantics~\citep{2022STEP}.
On the other hand, spatial-temporal positional encoding is designed to replace the standard positional encoding~\citep{2017Transformer,2021ViT} in Transformer. 
More importantly, it efficiently models the correlations among time series from the perspective of solving the indistinguishability of samples~\citep{2022STID, 2021STNorm, CANet}, 
avoiding the {\color{black}high complexity of conducting graph convolution~\citep{2019GWNet, 2018DCRNN} in the spatial dimension}.

Finally, we propose the training strategy: a two-stage strategy.
The first stage aims to train the hierarchical encoder based on the Mean Absolute Error~(MAE) between the intermediate prediction and ground truth.
In the second stage, we only train the decoder, while the parameters of the encoder are fixed to act as the multi-scale feature extractor.
The reason for adopting the two-stage strategy is that traffic forecasting tasks are different from tasks that employ an end-to-end strategy (e.g., semantic segmentation~\citep{2015UNet, 2021SwinUNet} and object detection~\citep{2017FPN} in computer vision).
Specifically, in computer vision tasks, pre-trained vision models (e.g., pre-trained ResNet~\citep{2016ResNet}) usually serve as the backbone to extract multi-scale features~\citep{2017FPN}.
However, there is no pre-trained model for time series that can extract multi-scale features.
Therefore, optimizing the feature extractor~(i.e., the encoder) and downstream networks~(i.e., the decoder) in an end-to-end fashion may be insufficient.
The experimental results in Section \ref{section:ablation} also verify this hypothesis.
Next, we introduce each component in detail.

\subsection{Input embedding}

\textbf{Segment embedding.}
Most previous works usually use single data points as the basic input units.
However, isolated points of time series usually give less semantics~\citep{2022STEP} and are more easily affected by noise.
Therefore, HUTFormer adopts segment embedding, i.e., dividing the input sequence into several segments to get the input tokens.
Specifically, given the time series $\mathbf{X}^i\in\mathbb{R}^{T \times C}$ from sensor $i$, HUTFormer divides it into $P$ non-overlapping segments of length $L$, i.e., $T=P*L$. We denote the $j$th segment as $\mathbf{X}^i_j\in\mathbb{R}^{LC}$.
Then, we conduct the input embedding layer based on these segments: 
\begin{equation}
\mathbf{S}_j^i=\mathbf{W}\cdot\mathbf{X}^i_j+\mathbf{b}
\end{equation}
where $\mathbf{S}_j^i\in\mathbb{R}^{d}$ is the embedding of segments $j$ of the time series from sensor $i$, and $d$ is the hidden dimension.
$\mathbf{W}\in\mathbb{R}^{d\times (LC)}$ and $\mathbf{b}\in\mathbb{R}^{d}$ are learnable parameters shared by all segments.

In summary, applying segment embedding brings two benefits.
First, it provides more robust semantics. 
Second, it significantly reduces the sequence length to reduce computational complexity.

\noindent
\textbf{Spatial-temporal positional encoding.}
In this study, we propose to replace the standard positional encoding in Transformer-based networks~\citep{2017Transformer, 2021ViT} with Spatial-Temporal Positional Encoding~(ST-PE).
Specifically, given the segment embedding $\mathbf{S}_j^i\in\mathbb{R}^{d}$ of segments $j$ from time series $i$, ST-PE conduct positional encoding on the spatial and temporal dimensions simultaneously:
\begin{equation}
    \mathbf{U}_j^i = \text{Linear}(\mathbf{S}_j^i \parallel \mathbf{E}^i \parallel \mathbf{T}_j^{TiD} \parallel \mathbf{T}_j^{DiW})
    \label{eq:stpe}
\end{equation}
On the spatial dimension, we define the spatial positional embeddings $\mathbf{E}\in\mathbb{R}^{N\times d_1}$, where $N$ is the number of time series~(i.e., sensors), and $d_1$ is the hidden dimension.
On the temporal dimension, we define two semantic positional embeddings,
$\mathbf{T}^{TiD}\in\mathbb{R}^{N_D\times d_2}$ and $\mathbf{T}^{DiW}\in\mathbb{R}^{N_W\times d_3}$, 
where $N_D$ is the number of time slots of a day~(determined by the sensor's sampling frequency) and $N_W=7$ is the number of days in a week. The temporal embeddings are thus shared among slots for the same time of the day and the same day of the week.
Semantic temporal positional embeddings are helpful since traffic systems usually reflect the periodicity of human society.
In addition, kindly note that all other baseline models~\citep{2018DCRNN, 2019GWNet, 2022D2STGNN, 2021AutoFormer, 2022FEDFormer, 2022Pyraformer} also use such temporal features, so there is no unfairness.
$\text{Linear}(\cdot)$ is a linear layer to reduce the hidden dimension.
$\mathbf{E}$, $\mathbf{T}^{TiD}$, and $\mathbf{T}^{DiW}$ are trainable parameters.

Embedding $\mathbf{E}$ is vital for reducing the complexity of modeling the spatial correlations between time series.
This is because attaching spatial embeddings plays a similar role to GCN in terms of solving the indistinguishability of samples~\citep{2022STID}, but with two primary advantages. 
On the one hand, it is more efficient than GCNs, which usually have $\mathcal{O}(N^2)$ complexity.
On the other hand, it does not generate many additional network parameters than approaches based on variable-specific modeling~\citep{2020AGCRN, Triformer}.

\subsection{Hierarchical encoder}
\begin{figure}[h]
  \centering
  \includegraphics[width=0.6\linewidth]{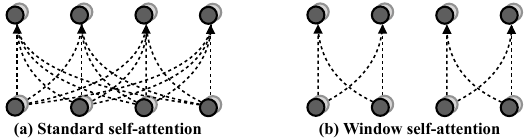}
  \caption{Standard self-attention v.s. window self-attention.}
  \label{fig:w-msa}
\end{figure}

\textbf{Window Transformer layer.}
Standard Transformer layers~\citep{2017Transformer} are designed based on the multi-head self-attention mechanism.
As shown in Fig. \ref{fig:w-msa}a, it computes the attention among all input tokens.
Therefore, each layer of the Transformer layer has an infinite receptive field, and many works~\citep{2021Informer, 2021AutoFormer, 2022FEDFormer, 2022Pyraformer} try to capture long-term dependencies based on such a feature.

However, the infinite receptive field makes the standard transformer layers unable to generate multi-scale features~\citep{2021SwinTransformer}.
Inspired by recent development in computer vision~\citep{2021SwinTransformer}, we apply the window self-attention in HUTFormer to extract the hierarchical multi-scale features.
An example of window self-attention with windows size 2 is shown in Fig. \ref{fig:w-msa}b.
Window self-attention forces calculating attention inside non-overlapping windows, thereby limiting the size of the receptive field.
By replacing multi-head self-attention in standard Transformer layers~\citep{2017Transformer} with the Window Multi-head Self-Attention~(W-MSA), we present the window Transformer layer:
\begin{equation}
    \begin{aligned}
    \mathbf{H}^{\mathrm{in}'} & = \text{W-MSA}(\text{LN}(\mathbf{H}^{\mathrm{in}})) + \mathbf{H}^{\mathrm{in}} \\
    \mathbf{H}^{\mathrm{out}} & = \text{MLP}(\text{LN}(\mathbf{H}^{\mathrm{in}'})) + \mathbf{H}^{\mathrm{in}'}
    \end{aligned}
\end{equation}
where $\text{LN}(\cdot)$ is the layer normalization, and $\text{MLP}(\cdot)$ is the multi-layer perceptron. $\mathbf{H}^{\mathrm{in}}\in\mathbb{R}^{P\times d}$ and $\mathbf{H}^{\mathrm{out}}\in\mathbb{R}^{P\times d}$ are the input and output sequences. $P$ is the sequence length, and $d$ is the hidden dimension.
By limiting the receptive field size, the window transformer layer is the basis for extracting multi-scale features.

\begin{figure}[h]
  \centering
  \includegraphics[width=0.6\linewidth]{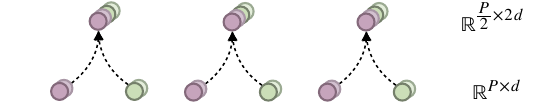}
  \caption{An illustration of segment merging.}
  \label{fig:segment_merging}
\end{figure}

\noindent
\textbf{Segment merging.} To generate hierarchical multi-scale representations, we adopt segment merging, which reduces the number of tokens and increases the number of hidden dimensions as the network gets deeper. 
As illustrated in Fig. \ref{fig:segment_merging}, segment merging divides the token series into non-overlapping groups of size 2, and concatenates the features within each group.

By combining the segment merging and window transformer layer, we get the basic block of the hierarchical encoder~(i.e., the blue block in Fig. \ref{fig:main_arch}).
Assuming $(\mathbf{H}^i)^{l}_{\mathrm{enc}}\in\mathbb{R}^{P^l\times d^l}$ is the representation of series $i$ after block $l$~($l\geq 1$) of the encoder, the $(l+1)$th block is computed as
\begin{equation}
    \begin{aligned}
    (\mathbf{H}^i)^{l'}_{\mathrm{enc}} &= \text{SegmentMerging}((\mathbf{H}^i)^{l}_{\mathrm{enc}})\\
    (\mathbf{H}^i)^{l+1}_{\mathrm{enc}} &= \text{WindowTransformer}((\mathbf{H}^i)^{(l')}_{\mathrm{enc}})
    \end{aligned}
\end{equation}
where $(\mathbf{H}^i)^{l+1}_{\mathrm{enc}}\in\mathbb{R}^{P^{l+1}\times d^{l+1}}$ is the representation of time series $i$ after block $l+1$ of the encoder. 
$P^{l+1}=\frac{P^l}{2}$ is the number of tokens after $(l+1)$th layer, and $d^{l+1}=2d^{l+1}$ is the hidden dimension.

\noindent
\textbf{Prediction layer.}
Assuming there are $S$ blocks in the encoder, HUTFormer makes an intermediate prediction with a linear layer:
\begin{equation}
    \hat{\mathbf{Y}}^i_{\mathrm{enc}} = \text{Linear}(\mathop{\parallel}\limits_{j=1}^{P^S}(\mathbf{H}^i_j)^{S}_{\mathrm{enc}})
\end{equation}
where $P^S$ is the number of tokens after the $S$th block.
$\hat{\mathbf{Y}}^i\in\mathbb{R}^{T_f\times C}$ is the prediction of time series $i$.
Considering the prediction from all $N$ time series, $\hat{\mathcal{Y}}^{enc}\in\mathbb{R}^{T_f\times N\times C}$, we compute the Mean Absolute Error~(MAE) as regression loss to train the hierarchical encoder:
\begin{equation}
    \mathcal{L}_{enc} = \frac{1}{T_f N C}\sum_{j=1}^{T_f}\sum_{i=1}^{N}\sum_{k=1}^{C}|\hat{\mathcal{Y}}_{ijk}^{enc} - \mathcal{Y}_{ijk}|
\end{equation}

\subsection{Hierarchical decoder}
Cross-scale Transformer layer.
The hierarchical decoder aims to effectively utilize the multi-scale features, to fine-tune each segment of the intermediate prediction.
However, as discussed in Section \ref{sec:methods:overview}, the history and future sequence in traffic forecasting tasks are not aligned, making the feature sequences extracted by the encoder and the decoder cannot be directly superimposed.
Therefore, we design a cross-scale attention mechanism to select and incorporate multi-scale features.
Different from self-attention, cross-scale attention utilizes the representations of the decoder as queries to retrieve the multi-scale features from the encoder.
For brevity, we denote $\mathbf{H}_{\mathrm{enc}}\in\mathbb{R}^{P_{\mathrm{enc}}\times d_{\mathrm{enc}}}$ as the representation from the encoder and $\mathbf{H}_{\mathrm{dec}}\in\mathbb{R}^{P_{\mathrm{dec}}\times d_{\mathrm{dec}}}$ as the corresponding representation from the decoder.
Then, the Cross-scale Attention~(CA) is computed as
\begin{equation}
    \begin{aligned}
    \text{CA}(\mathbf{H}_{\mathrm{enc}}, \mathbf{H}_{\mathrm{dec}})&=\text{Softmax}(\frac{\mathbf{H}_{\mathrm{dec}}(\mathbf{H}_{\mathrm{enc}}^{'})^T}{\sqrt{d_{\mathrm{dec}}}})\mathbf{H}_{\mathrm{enc}}^{'}\\
\mathbf{H}_{\mathrm{enc}}^{'}&=\text{Linear}(\mathbf{H}_{\mathrm{enc}})
    \end{aligned}
\end{equation}
The $\text{Linear}(\cdot)$ layer is used to transform the hidden dimension from $d_{\mathrm{enc}}$ to $d_{\mathrm{dec}}$.
By replacing the multi-head self-attention with Multi-head Cross-scale Attention~(MCA), we present the cross-scale Transformer layer as
\begin{equation}
    \begin{aligned}
    \mathbf{H}^{\mathrm{in}'}_{\mathrm{dec}} & = \text{MCA}(\text{LN}(\mathbf{H}^{\mathrm{in}}_{\mathrm{enc}}, \mathbf{H}^{\mathrm{in}}_{\mathrm{dec}})) + \mathbf{H}^{\mathrm{in}}_{\mathrm{dec}} \\
    \mathbf{H}^{\mathrm{out}}_{\mathrm{dec}} & = \text{MLP}(\text{LN}(\mathbf{H}^{\mathrm{in}'}_{\mathrm{dec}})) + \mathbf{H}^{\mathrm{in}'}_{\mathrm{dec}}
    \end{aligned}
\end{equation}
where $\mathbf{H}^{\mathrm{in}}_{\mathrm{enc}}$ is the multi-scale feature from the encoder, and $\mathbf{H}^{\mathrm{in}}_{\mathrm{dec}}$ is the input feature from the decoder.
$\mathbf{H}^{\mathrm{out}}_{\mathrm{dec}}$ is the output of the cross-scale Transformer layer.

\noindent
\textbf{Prediction layer.} Assuming $(\mathbf{H}^i_j)^S_{\mathrm{dec}}\in\mathbb{R}^{d_{\mathrm{dec}}}$ is the representation of the decoder's last block~(i.e., the $S$th block) for $j$th segment of $i$th time series, HUTFormer makes the final prediction for each segment with a shared linear layer:
\begin{equation}
    (\hat{\mathbf{Y}}^i_j)_{\mathrm{dec}} = \text{Linear}((\mathbf{H}^i_j)^{S}_{\mathrm{dec}})
\end{equation}
where $(\hat{\mathbf{Y}}^i_j)_{\mathrm{dec}}\in\mathbb{R}^{LC}$ is the final prediction of segment $j$ of time series $i$.
Similar to the encoder, we consider the prediction from all $P_{\mathrm{dec}}$ segments~($P_{\mathrm{dec}}\times L=T_f$) of all $N$ time series, $\hat{\mathcal{Y}}^{\mathrm{dec}}\in\mathbb{R}^{T_f\times N\times C}$, and compute the MAE loss to train the hierarchical decoder:
\begin{equation}
    \mathcal{L}_{\mathrm{dec}} = \frac{1}{T_f N C}\sum_{j=1}^{T_f}\sum_{i=1}^{N}\sum_{k=1}^{C}|\hat{\mathcal{Y}}_{ijk}^{\mathrm{dec}} - \mathcal{Y}_{ijk}|
\end{equation}
Kindly note that the parameters of the encoder are fixed during this stage to serve as a pre-training model for extracting robust hierarchical multi-scale representations of traffic data.
\section{Experiments}
\label{sec:exp}
In this section, we conduct extensive experiments on four real-world traffic datasets to validate the effectiveness of HUTFormer for long-term traffic forecasting.
First, we introduce the experimental settings, including datasets, baselines, and implementation details.
Then, we compare HUTFormer with other state-of-the-art traffic forecasting baselines and long-sequence time series forecasting baselines.
Furthermore, we conduct more experiments to evaluate the impact of important components and strategies, including the effectiveness of the hierarchical U-net structure, the input embedding strategy, and the two-stage training strategy.

\subsection{Experimental setting}
\noindent
\textbf{Datasets.}
We conduct experiments on four commonly used traffic datasets, including two traffic speed datasets~(METR-LA and PEMS-BAY) and two traffic flow datasets~(PEMS04 and PEMS08).
The statistical information is summarized in Table \ref{tab:datasets}.

\begin{itemize}
    \item METR-LA is a traffic speed dataset collected from loop-detectors located on the LA County road network~\citep{METR-LA}. 
    It contains data of 207 selected sensors over a period of 4 months from Mar to Jun in 2012~\citep{2018DCRNN}. 
    The traffic information is recorded at the rate of every 5 min, and the total number of time slices is 34,272.
    \item PEMS-BAY is a traffic speed dataset collected from California Transportation Agencies (CalTrans) Performance Measurement System (PeMS)~\citep{PEMS-BAY}.
    It contains data of 325 sensors in the Bay Area over a period of 6 months from Jan 1st 2017 to May 31th 2017~\citep{2018DCRNN}.
    The traffic information is recorded at the rate of every 5 min, and the total number of time slices is 52,116.
    \item PEMS04 is a traffic flow dataset also collected from CalTrans PeMS.
    It contains data of 307 sensors over a period of 2 months from Jan 1st 2018 to Feb 28th 2018~\citep{2019ASTGCN}.
    The traffic information is recorded at the rate of every 5 min, and the total number of time slices is 16,992.
    \item \color{black}{PEMS08 is a public traffic flow dataset collected from CalTrans PeMS. 
    Specifically, PEMS08 contains data of 170 sensors in San Bernardino over a period of 2 months from July 1st 2018 to Aug 31th 2018~\citep{2019ASTGCN}.
    The traffic information is recorded at the rate of every 5 min, and the total number of time slices is 17,856.}
    \item \color{black}{ETTh1, ETTm1, and Weather} are non-traffic time series datasets used to verify the generalization of the proposed HUTFormer. Due to space limitations, we omit their details. Interested readers can refer to~\citep{2021Informer}.
\end{itemize}

\begin{table}
\caption{Statistics of datasets.}
\label{tab:datasets}
\begin{center}
  \begin{tabular}{c|c|c|c|c}
    \toprule
    Type & Dataset & \# Sample & \# Sensor & Sample rate\\
    \midrule
    \multirow{2}*{Speed} &
    METR-LA & 34,272  & 207 & 5 min\\
    & PEMS-BAY & 52,116  & 325 & 5 min\\
    \midrule
    \multirow{2}*{Flow}  & 
    PEMS04 & 16,992 & 307 & 5 min\\
    & PEMS08 & 17,856 & 170 & 5 min\\
    \bottomrule
  \end{tabular}
\end{center}
\end{table}

\begin{figure}[htbp]
  \centering
  \includegraphics[width=0.9\linewidth]{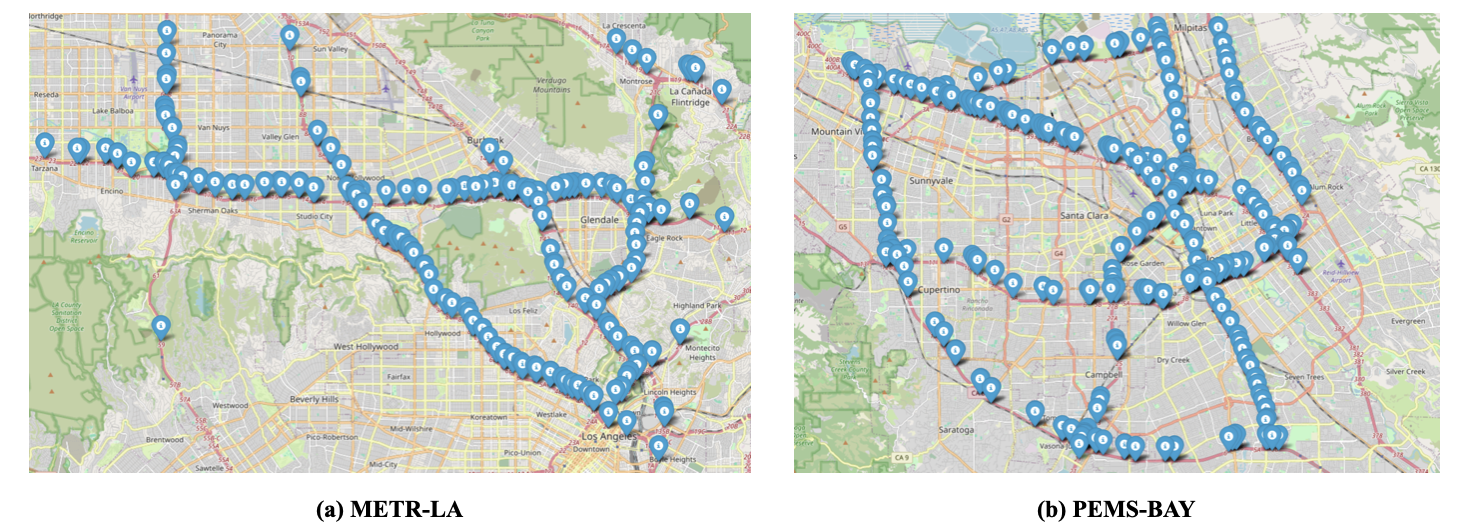}
  \caption{
  Spatial typologies of METR-LA and PEMS-BAY datasets.}
  \label{figure:spatial_typologies}
\end{figure}

{\color{black}
Spatial attributes—especially road-network topology—are indispensable to traffic forecasting tasks. The following describes in detail the spatial information used by each traffic dataset\footnote{ETTh1, ETTm1, and Weather are not traffic time-series datasets; they are non-spatiotemporal benchmarks used to evaluate the generalization capability of HUTFormer, and are therefore omitted here.}.
\begin{itemize}
    \item METR-LA and PEMS-BAY: These datasets comprise 207 and 325 sensor nodes, respectively. Their geographic distributions are illustrated in Fig.~\ref{figure:spatial_typologies}. Following prior work~\citep{2018DCRNN}, the spatial adjacency matrix $A$ is computed with a thresholded Gaussian kernel~\citep{shuman2013emerging}:
    \begin{equation}
        A_{ij} =
            \begin{cases}
            \exp\!\left(-\dfrac{\operatorname{dist}(v_i,v_j)^{2}}{\sigma^{2}}\right), & \operatorname{dist}(v_i,v_j) \le \kappa \\[6pt]
            0, & \text{otherwise}
            \end{cases}
    \end{equation}
    where $\operatorname{dist}(v_i,v_j)$ denotes the road-network distance between nodes $v_i$ and $v_j$, $\sigma$ is the standard deviation of all pairwise distances, and $\kappa$ is the distance threshold.
    \item PEMS04 and PEMS08: These datasets contain 307 and 170 nodes, respectively. Because the original releases \citep{2018STGCN,2019ASTGCN} only contain the distance between sensors without the raw latitude–longitude coordinates, we omit their spatial visualizations. Consistent with the previous studies~\citep{2018STGCN,2019ASTGCN}, the spatial adjacency matrix $A$ is defined as
    \begin{equation}
        A_{ij} =
            \begin{cases}
            1, & \operatorname{dist}(v_i,v_j) \le 3.5\ \text{miles} \\[6pt]
            0, & \text{otherwise}
            \end{cases}
    \end{equation}
\end{itemize}
}

\noindent
Baselines.
On the one hand, 
we select six traffic forecasting baselines, including: 
\begin{itemize}
    \item DCRNN~\citep{2018DCRNN} is one of the earliest works for STGNN-based traffic forecasting, which replaces the fully connected layer in GRU~\citep{2014GRU} by diffusion convolutional layer to form a Diffusion Convolutional Gated Recurrent Unit. 
    \item Graph WaveNet~\citep{2019GWNet} is a traffic forecasting model, which stacks gated temporal convolutional layer and GCN layer by layer to jointly capture the spatial and temporal dependencies.
    \item MTGNN~\citep{2020MTGNN} is a traffic forecasting model, which extends Graph WaveNet through the mix-hop propagation layer in the spatial module, the dilated inception layer in the temporal module, and a delicate graph learning layer.
    \item STID~\citep{2022STID} is a simple but effective baseline for traffic forecasting, which identifies the indistinguishability of samples in both spatial and temporal dimensions as a key bottleneck, and addresses the indistinguishability by attaching spatial and temporal identities.
    \item STEP~\citep{2022STEP} is a traffic forecasting model, which enhances existing STGNNs with the help of a time series pre-training model.
    It significantly extends the length of historical data.
    \item D$^2$STGNN~\citep{2022D2STGNN} is a state-of-the-art traffic forecasting model, which identifies the diffusion process and inherent process in traffic data, and further decouples them for better modeling. 
\end{itemize}

On the other hand, 
we also select six long-sequence forecasting baselines, 
including: 
\begin{itemize}
    \item HI~\citep{2020HI} is a basic baseline for long-sequence time series forecasting problems, which  directly takes the most recent time steps in the input as output.
    \item DLinear~\citep{2023DLinear} is a simple but effective long-sequence time series forecasting model, which decomposes the time series into a trend and a remainder series and employs two one-layer linear networks to model these two series.
    \item Informer~\citep{2021Informer} is a model for long-sequence time series forecasting, which designs a ProbSparse self-attention mechanism and distilling operation to handle the challenges of the quadratic complexity in the standard Transformer.
    Also, it carefully designs a generative decoder to alleviate the limitation of standard encoder-decoder architecture.
    \item Autoformer~\citep{2021AutoFormer} is a model for long-sequence time series forecasting, which is proposed as a decomposition architecture by embedding the series decomposition block as an inner operator. Besides, it designs an efficient Auto-Correlation mechanism to conduct dependencies discovery and information aggregation at the series level.
    \item FEDformer~\citep{2022FEDFormer} is a frequency-enhance Transformer for long-sequence time series forecasting. It proposes an attention mechanism with low-rank approximation in frequency and a mixture of experts decomposition to control the distribution shifting.
    \item Pyraformer~\citep{2022Pyraformer} is a pyramidal attention-based model for long-sequence time series forecasting.
    Pyramidal attention can effectively describe short and long temporal dependencies.
    \item \color{black}Crossformer~\citep{2023Crossformer} is a Transformer-based model utilizing cross-dimension dependency for multivariate time-series (MTS) forecasting.
    \item \color{black}PatchTST~\citep{2023PatchTST} proposes an effective design of Transformer-based models for time series forecasting tasks by introducing two key components: Patching and channel-independent structure.
    
\end{itemize}

\noindent
\textbf{Metrics.}
In this study, we evaluate the performances of all baselines by Mean Absolute Error (MAE) and Mean Absolute Percentage Error (MAPE) metrics. 
First, the MAE metric reflects the absolute prediction error, but is affected by the units of the dataset. 
For example, traffic speed datasets usually take values between 0 and 70 km/h, while traffic flow datasets usually take values between zero and hundreds.
Thus, we also adopt MAPE, which can eliminate the impact of data units and reflects the relative error, helping to understand the accuracy more intuitively.

\begin{table}[t]
\caption{Optimization settings.}
\label{tab:opt_enc}
\centering  
\begin{tabular}{p{3cm}|p{4cm}}
\toprule
Config  & Value \\
\midrule
Optimizer & Adam\\
Learning rate & 0.0005\\
Batch size & 64\\
Weight decay & 0.0001\\
Learning rate schedule & MultiStepLR\\
Milestones & [1, 40, 80, 120]\\
Gamma & 0.5\\
Gradient clip & 5\\
\bottomrule
\end{tabular}
\end{table}

\noindent
\textbf{Implementation.}
For all datasets, we use historical $T_p=288$ time steps~(i.e., 1 day) to predict future $T_f=288$ time steps.
For HUTFormer, we set the segment length $L$ to 12, and the number of segments $P=24$~($L\times P=288)$.
We set the window size to $3$.
We set the hidden dimension of temporal embedding $\textbf{T}^{TiD}$ to 8, while others $d$ to 32.
The depth of HUTFormer is set to $4$.
For baselines, we adopt the default settings.
Moreover, as discussed before, STGNNs can not directly handle the long-term traffic forecasting task due to their high complexity.
Therefore, we first apply the segment embedding to reduce the length of input tokens for them\footnote{
Methods implemented with segment embeddings are marked with $*$.}.
{\color{black}On the one hand, all baseline models are based on their official open-source code, with only modifications to the model input (introducing segment embedding technology), and we ensure that both HUTFormer and the baseline models use the same hyperparameters (segment size and stride size), thereby maintaining fairness in model structure. On the other hand, all models are trained using a unified and scalable pipeline~\citep{BasicTS}, ensuring fairness in the training process. }

\noindent {Optimization settings.}
For both encoding and decoding stages, we apply the optimization settings in Tabel \ref{tab:opt_enc}.
Specifically, we adopt Adam as our optimizer, and set learning rate and weight decay to $0.0005$ and $0.0001$, respectively.
The batch size is set to 64.
In addition, we use a learning rate scheduler, MultiStepLR, which adjusts the learning rate at epochs 1, 40, 80, and 120 with gamma 0.5.
Moreover, the gradient clip is set to 5.
All the experiments in Section \ref{sec:exp} are running on an Intel(R) Xeon(R) Gold 5217 CPU @ 3.00GHz, 128G RAM computing server, equipped with RTX 3090 graphics cards.

\begin{table*}[htpb]
    \centering
    \renewcommand\arraystretch{1}
    \caption{Long-term traffic forecasting on traffic speed datasets METR-LA and PEMS-BAY. 
    }
    \label{tab:traffic_speed}
    \scalebox{0.8}{
    \begin{tabular}{cccc|cc|cc|cc|cc|cc}
      \toprule
      \midrule
      \multirow{2}*{Data} &
      \multirow{2}*{Method} & 
      \multicolumn{2}{c}{@Horizon 12}& 
      \multicolumn{2}{c}{@Horizon 48}& 
      \multicolumn{2}{c}{@Horizon 96}& 
      \multicolumn{2}{c}{@Horizon 144}& 
      \multicolumn{2}{c}{@Horizon 192}& 
      \multicolumn{2}{c}{@Horizon 288} \\
      \cmidrule(r){3-4} \cmidrule(r){5-6} \cmidrule(r){7-8} \cmidrule(r){9-10} \cmidrule(r){11-12} \cmidrule(r){13-14}
      &  & MAE & MAPE (\%) & MAE & MAPE (\%) & MAE & MAPE (\%) & MAE & MAPE (\%) & MAE & MAPE (\%) & MAE & MAPE(\%)\\
      \midrule
    \midrule
    \multirow{12}*{\rotatebox{90}{METR-LA}}
      & HI & 10.44 & 23.21 & 10.42 & 23.19 & 10.43 & 23.23 & 10.43 & 23.32 & 10.40 & 23.34 & 10.22 & 22.81\\
      & DLinear & 7.61 & 16.19 & 12.86 & 23.79 & 12.99 & 23.11 & 12.90 & 23.48 & 12.89 & 23.15 & 13.07 & 23.33\\
      & Informer & 4.65 & 15.52 & 4.86 & 16.54 & 4.98 & 17.16 & 5.07 & 17.41 & 5.07 & 17.30 & 5.06 & 17.14\\
      & Autoformer & 7.23 & 19.25 & 7.27 & 19.73 & 7.45 & 20.23 & 7.83 & 21.49 & 7.74 & 20.98 & 8.41 & 22.43\\
      & FEDformer & 8.78 & 22.29 & 9.11 & 22.69 & 9.12 & 22.75 & 9.54 & 24.18 & 9.81 & 24.76 & 10.13 & 25.56\\
      & Pyraformer & 4.22 & 12.84 & 4.55 & 14.93 & 4.75 & 15.81 & 4.80 & 15.89 & 4.81 & 15.68 & 4.62 & 14.79\\
      & PatchTST & 4.43 & 13.58 & 5.02 & 16.37 & 5.14 & 16.64 & 5.19 & 16.98 & 5.21 & 16.67 & 5.25 & 17.16 \\
      \cmidrule{2-14}
      & DCRNN$^*$ & 4.07 & 12.74 & 4.39 & 14.08 & 4.44 & 14.02 & 4.46 & 14.16  & 4.51 & 14.41 & 4.71 & 15.59\\
      & GWNet$^*$ & 3.87 & 12.18 & 4.19 & 13.60 & 4.25 & 13.62 & 4.42 & 14.56 & 4.58 & 15.40 & 4.51 & 15.09\\
      & MTGNN$^*$ & 4.01 & 12.31 & 4.31 & 13.84 & 4.53 & 14.85 & 4.59 & 14.77 & 4.57 & 15.18 & 4.75 & 15.93\\
      & STID & 3.84 & 12.17 & 4.13 & 14.11 & 4.04 & 13.05 & 4.11 & 13.65 & 4.15 & 14.07 & 4.17 & 13.83\\
      & STEP$^*$ & 3.74 & 11.60 & 4.14 & 13.24 & 4.22 & 13.52 & 4.38 & 14.07 & 4.34 & 13.96 & 4.43 & 14.42\\
      & D$^2$STGNN$^*$ & 3.71 & 11.24 & 3.96 & 12.84 & 3.99 & 13.26 & 4.05 & 13.17 & 4.05 & 13.36 & 4.09 & 12.78\\
      \cmidrule{2-14}
      & HUTFormer & \textbf{3.59} & \textbf{10.93} & \textbf{3.77} & \textbf{11.88} & \textbf{3.79} & \textbf{11.86} & \textbf{3.80} & \textbf{12.08} & \textbf{3.82} & \textbf{12.18} & \textbf{3.84} & \textbf{12.28}\\
    \midrule
    \midrule
    \multirow{12}*{\rotatebox{90}{PEMS-BAY}}
      & HI & 3.37 & 7.84 & 3.36 & 7.80 & 3.36 & 7.77 & 3.36 & 7.76 & 3.36 & 7.74 & 3.38 & 7.79\\
      & DLinear & 2.70 & 6.28 & 3.14 & 7.75 & 3.13 & 7.77 & 3.15 & 7.76 & 3.15 & 7.78 & 3.23 & 7.90\\
      & Informer & 2.77 & 6.65 & 2.80 & 6.88 & 2.84 & 7.06 & 2.83 & 7.07 & 2.82 & 6.98 & 2.92 & 7.16\\
      & Autoformer & 3.15 & 7.48 & 3.24 & 7.85 & 3.30 & 8.00 & 3.37 & 8.10 & 3.39 & 8.15 & 4.35 & 11.25\\
      & FEDformer & 3.04 & 7.55 & 3.14 & 7.61 & 3.13 & 7.58 & 3.32 & 8.00 & 3.42 & 8.45 & 3.67 & 9.33\\
      & Pyraformer & 2.53 & 6.21 & 2.71 & 6.72 & 2.64 & 6.39 & 2.74 & 6.65 & 2.75 & 6.68 & 2.77 & 6.81\\
      & PatchTST & 2.35 & 5.94 & 2.92 & 7.45 & 2.96 & 7.52 & 3.00 & 7.62 & 3.01 & 7.67 & 3.10 & 7.73\\
      \cmidrule{2-14}
      & DCRNN$^*$ & 2.18 & 5.49 & 2.52 & 6.49 & 2.54 & 6.43 & 2.66 & 6.79 & 2.67 & 6.80 & 2.66 & 6.62\\
      & GWNet$^*$ & 2.01 & 5.11 & 2.35 & 5.91 & 2.40 & 5.98 & 2.47 & 6.35 & 2.46 & 6.24 & 2.46 & 6.09\\
      & MTGNN$^*$ & 2.17 & 5.40 & 2.45 & 6.11 & 2.51 & 6.04 & 2.52 & 6.13 & 2.57& 6.19 & 2.70 & 6.40\\
      & STID & 2.02 & 5.02 & 2.29 & 5.66 & 2.32 & 5.69 & 2.33 & 5.72 & 2.32 & 5.67 & 2.38 & 5.81\\
      & STEP$^*$ & 2.00 & 4.94 & 2.33 & 5.93 & 2.38 & 6.05 & 2.44 & 6.26 & 2.45 & 6.24 & 2.54 & 6.41\\
      & D$^2$STGNN$^*$ & 2.04 & 4.97 & 2.26 & 5.44 & 2.29 & 5.60 & 2.34 & 5.55 & 2.31 & 5.50 & 2.38 & 5.64\\
      \cmidrule{2-14}
      & HUTFormer & \textbf{1.93} & \textbf{4.62} & \textbf{2.18} & \textbf{5.16} & \textbf{2.21} & \textbf{5.24} & \textbf{2.22} & \textbf{5.24} & \textbf{2.23} & \textbf{5.25} & \textbf{2.28} & \textbf{5.35}\\
    \midrule
    \toprule
    \end{tabular}
    }
  \end{table*}

\begin{table*}[htpb]
\renewcommand\arraystretch{1}
    \centering
    \caption{Long-term traffic forecasting on traffic flow datasets PEMS04 and PEMS08. 
    }
    \label{tab:traffic_flow}
    \scalebox{0.8}{
    \begin{tabular}{cccc|cc|cc|cc|cc|cc}
      \toprule
      \midrule
      \multirow{2}*{Data} &
      \multirow{2}*{Method} & 
      \multicolumn{2}{c}{@Horizon 12}& 
      \multicolumn{2}{c}{@Horizon 48}& 
      \multicolumn{2}{c}{@Horizon 96}& 
      \multicolumn{2}{c}{@Horizon 144}& 
      \multicolumn{2}{c}{@Horizon 192}& 
      \multicolumn{2}{c}{@Horizon 288} \\
      \cmidrule(r){3-4} \cmidrule(r){5-6} \cmidrule(r){7-8} \cmidrule(r){9-10} \cmidrule(r){11-12} \cmidrule(r){13-14}
      &  & MAE & MAPE (\%) & MAE & MAPE (\%) & MAE & MAPE (\%) & MAE & MAPE (\%) & MAE & MAPE (\%) & MAE & MAPE(\%)\\
      \midrule
    \midrule
    \multirow{12}*{\rotatebox{90}{PEMS04}}
      & HI & 41.73 & 28.46 & 41.16 & 28.61 & 41.38 & 28.62 & 41.28 & 28.42 & 30.99 & 27.34 & 39.58 & 26.49\\
      & DLinear & 27.29 & 19.83 & 37.20 & 26.51 & 37.50 & 26.78 & 37.57 & 26.87 & 37.17 & 25.27 & 36.87 & 25.21\\
      & Informer & 25.94 & 17.56 & 25.72 & 18.05 & 25.60 & 18.27 & 25.98 & 17.81 & 26.42 & 17.67 & 27.42 & 18.57\\
      & Autoformer & 29.94 & 28.00 & 31.30 & 27.41 & 31.47 & 27.73 & 31.95 & 27.89 & 32.03 & 28.03 & 33.34 & 29.82\\
      & FEDformer & 34.94 & 34.33 & 32.24 & 37.23 & 33.90 & 34.33 & 35.12 & 41.26 & 35.16 & 34.08 & 41.83 & 51.01\\
      & Pyraformer & 23.40 & 17.18 & 25.40 & 18.80 & 26.45 & 19.89 & 26.22 & 19.01 & 26.51 & 19.18 & 26.58 & 20.57\\
      & PatchTST & 22.75 & 16.67 & 29.37 & 21.85 & 30.63 & 23.15 & 32.01 & 24.00 & 30.54 & 21.54 & 31.50 & 24.00 \\
      \cmidrule{2-14}
      & DCRNN$^*$ & 22.25 & 16.59 & 24.42 & 18.89 & 25.20 & 19.17 & 26.31 & 19.61 & 27.32 & 19.74 & 28.04 & 21.02\\
      & GWNet$^*$ & 22.24 & 16.51 & 23.50 & 18.29 & 24.08 & 18.07 & 24.85 & 18.21 & 25.83 & 18.98 & 31.17 & 21.00\\
      & MTGNN$^*$ & 21.75 & 15.93 & 23.04 & 17.81 & 24.33 & 17.80 & 25.56 & 17.68 & 25.80 & 17.85 & 26.78 & 20.64\\
      & STID & 21.01 & 15.24 & 22.77 & 16.61 & 23.39 & 16.87 & 24.06 & 17.08 & 24.43 & 17.22 & 25.19 & 17.49\\
      & STEP$^*$ & 20.82 & 15.56 & 22.23 & 17.11 & 22.87 & 17.21 & 24.46 & 17.97 & 24.89 & 17.40 & 26.18 & 18.47\\
      & D$^2$STGNN$^*$ & 21.55 & 16.03 & 22.98 & 17.04 & 24.16 & 17.57 & 24.50 & 17.93 & 24.59 & 17.19 & 24.79 & 17.97\\
      \cmidrule{2-14}
      & HUTFormer & \textbf{19.61} & \textbf{13.59} & \textbf{21.54} & \textbf{14.95} & \textbf{21.96} & \textbf{15.22} & \textbf{22.66} & \textbf{15.30} & \textbf{23.10} & \textbf{15.35} & \textbf{23.43} & \textbf{15.71}\\
    \midrule
    \midrule
    \multirow{12}*{\rotatebox{90}{PEMS08}}
      & HI & 37.33 & 25.01 & 37.31 & 25.07 & 37.23 & 25.05 & 37.09 & 25.02 & 36.94 & 24.98 & 36.40 & 24.76\\
      & DLinear & 22.91 & 17.23 & 34.13 & 24.15 & 34.34 & 25.54 & 34.44 & 23.80 & 34.52 & 23.91 & 35.11 & 23.71\\
      & Informer & 24.55 & 14.76 & 24.80 & 15.03 & 24.72 & 15.03 & 25.07 & 15.11 & 24.82 & 14.91 & 25.09 & 15.61\\
      & Autoformer & 31.36 & 25.44 & 32.29 & 27.13 & 33.19 & 27.45 & 32.98 & 26.15 & 33.57 & 25.78 & 36.75 & 28.82\\
      & FEDformer & 24.62 & 20.01 & 26.76 & 21.85 & 28.56 & 23.02 & 30.33 & 24.47 & 29.11 & 23.14 & 29.91 & 24.47\\
      & Pyraformer & 21.92 & 14.43 & 23.00 & 14.70 & 23.80 & 15.46 & 24.45 & 16.88 & 24.34 & 16.17 & 22.71 & 14.79\\
      & PatchTST & 16.94 & 11.37 & 21.27 & 15.10 & 22.56 & 16.39 & 23.22 & 17.40 & 23.18 & 17.70 & 23.73 & 17.35 \\
      \cmidrule{2-14}
      & DCRNN$^*$ & 18.64 & 13.47 & 20.42 & 14.92 & 20.97 & 15.11 & 21.63 & 15.51 & 22.45 & 16.23 & 22.95 & 16.72\\
      & GWNet$^*$ & 17.07 & 11.57 & 19.55 & 11.93 & 20.38 & 14.33 & 20.49 & 14.82 & 20.00 & 14.68 & 20.29 & 15.20\\
      & MTGNN$^*$ & 17.75 & 12.61 &19.27 & 13.35 & 19.99 & 13.85 & 20.68 & 15.00 & 20.95 & 14.65 & 22.16 & 15.68\\
      & STID & 16.40 & 11.42 & 18.53 & 13.26 & 19.17 & 13.66 & 19.59 & 13.78 & 19.59 & 14.03 & 20.23 & 15.35\\
      & STEP$^*$ & 16.67 & 11.34 & 19.05 & 14.00 & 19.74 & 14.74 & 20.15 & 14.88 & 19.80 & 14.84 & 20.37 & 15.54\\
      & D$^2$STGNN$^*$ & 17.27 & 11.47 & 18.45 & 12.35 & 18.97 & 12.63 & 19.33 & 12.81 & 19.09 & 12.34 & 19.55 & 12.93\\
      \cmidrule{2-14}
      & HUTFormer & \textbf{15.18} & \textbf{10.09} & \textbf{16.72} & \textbf{11.26} & \textbf{17.23} & \textbf{11.55} & \textbf{17.59} & \textbf{11.74} & \textbf{17.83} & \textbf{11.84} & \textbf{18.44} & \textbf{12.20}\\
    \midrule
    \toprule
    \end{tabular}
    }
  \end{table*}

\subsection{Main results}

\label{section:main_results}
\noindent
\textbf{Settings.}
We follow the dataset division in previous works.
Specifically, for traffic speed datasets (METR-LA and PEMS-BAY), we use 70\%, 10\%, and 20\% of the data for training, validating, and testing, respectively.
For traffic flow datasets (PEMS04 and PEMS08), we use 60\%, 20\%, and 20\% of data for training, validating, and testing, respectively.
We compare the performance at 1, 4, 8, 12, 16, and 24 hours~(horizon 12, 48, 96, 144, 192, and 288) of forecasting on the MAE and MAPE metrics.

\noindent
\textbf{Results.}
The results of traffic speed and flow forecasting are shown in Table \ref{tab:traffic_speed} and \ref{tab:traffic_flow}, respectively.
In general, HUTFormer consistently outperforms all baselines, indicating its effectiveness.
{\color{black}Notably, Crossformer~\citep{2023Crossformer} suffers from out-of-memory issues due to its high complexity and is therefore ignored in Table \ref{tab:traffic_speed} and \ref{tab:traffic_flow}.}

Long-sequence forecasting models do not perform well on traffic forecasting tasks.
We conjecture that the main reason is that these models do not fit the characteristics of traffic data.
First, there exist strong correlations between the time series of traffic data.
For example, due to the constraint of road networks, time series from adjacent sensors or from similar geographical functional areas may be more similar~\citep{2019STMetaNet}.
Understanding and exploiting the correlations between time series is essential for traffic forecasting.
However, long-sequence forecasting models are usually not concerned with such spatial dependencies.
Second, as discussed in Section \ref{section:intro}, the long-term traffic forecasting task requires exploiting multi-scale representations to capture the complex dynamics of traffic data. 
However, most long-term sequence forecasting models mainly focus on capturing global dependencies based on self-attention mechanisms.
For example, Informer~\citep{2021Informer} optimizes the efficiency of the original self-attention mechanism through the ProbSparse mechanism.
Autoformer~\citep{2021AutoFormer} conducts the dependencies discovery at the series level.
They can not generate and utilize multi-scale representations of traffic data.
In summary, the above-mentioned uniqueness of long-term traffic forecasting tasks significantly affects the effectiveness of long-sequence forecasting models.

Compared to long-sequence forecasting models, 
traffic forecasting models achieve better performance.
This is mainly because they model correlations between time series with the help of graph convolution.
Most of them~\citep{2018DCRNN,2019GWNet,2020MTGNN,2022STEP, 2022D2STGNN} utilize diffusion convolution, a variant of graph convolution, to model the diffusion process at each time step.
However, there is no free lunch.
The graph convolution brings a high complexity~\citep{2022STEP}.
As mentioned earlier, we had to implement these models with the segment embedding in HUTFormer to reduce the length of input tokens to make them runnable.
Kindly note that although the latest baseline STEP~\citep{2022STEP} can handle long-term historical data, it still requires a downstream STGNN as the backend, which can only make short-term future predictions.
In summary, these models only focus on short-term traffic forecasting and do not consider the uniqueness of long-term traffic forecasting, i.e., exploiting multi-scale representations.

Compared to all baselines, HUTFormer achieves state-of-the-art performances by sufficiently addressing the issues of long-term traffic forecasting tasks.
Specifically, on the one hand, HUTFormer efficiently handles the correlations between long-term time series with spatial-temporal positional encoding and segment embedding.
On the other hand, HUTFormer effectively generates and utilizes multi-scale representations based on the hierarchical U-net.

\subsection{Efficiency}
\label{appendix:efficiency}

In this section, we conduct more experiments to evaluate the efficiency of the HUTFormer variants in Section \ref{section:ablation}.
We conduct experiments with a single NVIDIA V100 graphics card with 32 GB memory, and report the GPU memory usage and running time.
Specifically, for the two-stage training variants, we report the largest GPU memory usage of the two stages and report the sum of the running time in the encoding and decoding stages.
We conduct experiments on the METR-LA dataset.

\begin{figure}[htbp]
  \centering
  \includegraphics[width=0.7\linewidth]{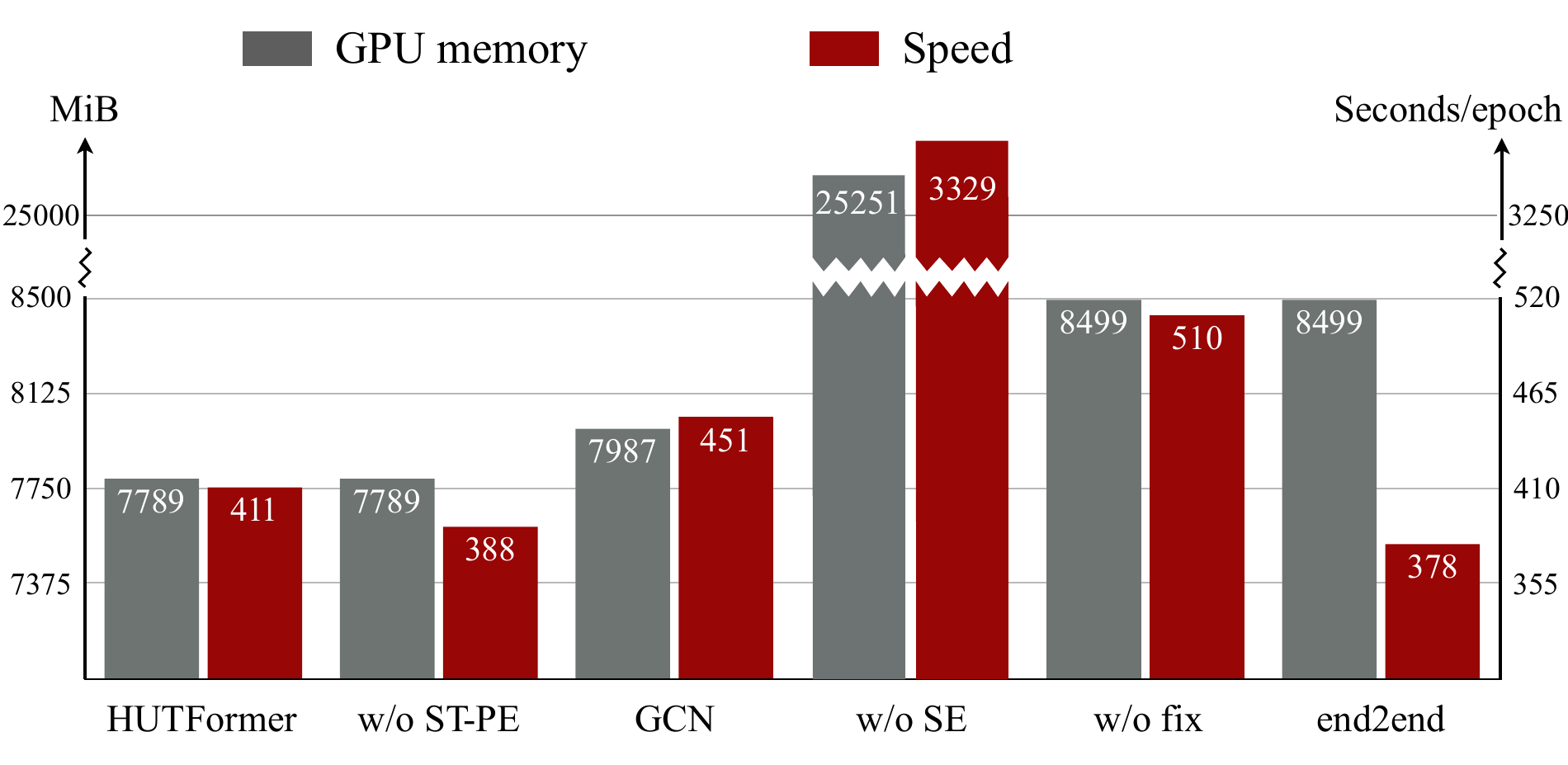}
  \caption{
  Efficiency study.}
  \label{figure:appendix_efficency}
\end{figure}

The results are shown in Fig. \ref{figure:appendix_efficency}.
First, we can see that removing the segment embedding~(i.e., w/o SE) will significantly increase the computational complexity, and require more GPU memory.
Second, compared with applying GCN, HUTFormer is more efficient and effective by leveraging the spatial-temporal positional encoding, which does not increase much complexity.

\subsection{Generalization}

\begin{table*}
    \centering
    \renewcommand\arraystretch{1}
    \caption{Experiments on ETTh1, ETTm1, and Weather datasets. 
    }
    \label{tab:more_datasets}
    \scalebox{0.9}{
    \begin{tabular}{cccc|cc|cc|cc|cc|cc}
      \toprule
      \midrule
      \multirow{2}*{Data} &
      \multirow{2}*{Method} & 
      \multicolumn{2}{c}{@Horizon 12}& 
      \multicolumn{2}{c}{@Horizon 48}& 
      \multicolumn{2}{c}{@Horizon 96}& 
      \multicolumn{2}{c}{@Horizon 144}& 
      \multicolumn{2}{c}{@Horizon 192}& 
      \multicolumn{2}{c}{@Horizon 288} \\
      \cmidrule(r){3-4} \cmidrule(r){5-6} \cmidrule(r){7-8} \cmidrule(r){9-10} \cmidrule(r){11-12} \cmidrule(r){13-14}
      &  & MAE & MSE & MAE & MSE & MAE & MSE & MAE & MSE & MAE & MSE & MAE & MSE\\
      \midrule
      \midrule
    \multirow{7}*{\rotatebox{90}{ETTh1}}
& Informer & 0.62 & 0.82 & 0.69 & 0.91 & 0.82 & 1.10 & 0.90 & 1.25 & 0.96 & 1.43 & 0.85 & 1.17\\
& Autoformer & 0.45 & 0.44 & 0.47 & 0.47 & 0.48 & 0.50 & 0.48 & 0.52 & 0.50 & 0.53 & 0.51 & 0.53\\
& FEDformer & 0.42 & 0.37 & 0.43 & 0.40 & 0.44 & 0.43 & 0.46 & 0.46 & 0.46 & 0.47 & 0.52 & 0.57\\
& Pyraformer & 0.56 & 0.63 & 0.57 & 0.64 & 0.65 & 0.79 & 0.74 & 0.92 & 0.76 & 1.03 & 0.79 & 1.09\\
& Triformer & 0.44 & 0.44 & 0.46 & 0.48 & 0.48 & 0.52 & 0.49 & 0.55 & 0.50 & 0.55 & 0.51 & 0.56\\
& Crossformer & 0.39 & 0.35 & 0.40 & 0.38 & 0.44 & 0.44 & 0.45 & 0.46 & 0.45 & 0.47 & 0.48 & 0.49\\
& PatchTST & 0.37 & 0.32 & \textbf{0.38} & \textbf{0.35} & 0.42 & 0.42 & \textbf{0.43} & 0.46 & \textbf{0.45} & \textbf{0.47} & 0.48 & 0.49\\
      \cmidrule{2-14}
      & HUTFormer & \textbf{0.36} & \textbf{0.31} & \textbf{0.38} & \textbf{0.35} & \textbf{0.41} & \textbf{0.41} & \textbf{0.43} & \textbf{0.44} & \textbf{0.45} & \textbf{0.47} & \textbf{0.47} & \textbf{0.47}\\
    \midrule
    \midrule
    \multirow{7}*{\rotatebox{90}{ETTm1}}
& Informer & 0.53 & 0.59 & 0.60 & 0.67 & 0.63 & 0.74 & 0.68 & 0.85 & 0.72 & 0.91 & 0.74 & 0.95\\
& Autoformer & 0.49 & 0.53 & 0.53 & 0.61 & 0.53 & 0.62 & 0.54 & 0.63 & 0.54 & 0.63 & 0.62 & 0.76\\
& FEDformer & 0.37 & 0.29 & 0.41 & 0.37 & 0.43 & 0.40 & 0.44 & 0.42 & 0.43 & 0.42 & 0.46 & 0.47\\
& Pyraformer & 0.52 & 0.53 & 0.64 & 0.80 & 0.62 & 0.71 & 0.71 & 0.89 & 0.59 & 0.65 & 0.71 & 0.88\\
& Triformer & 0.34 & 0.26 & 0.39 & 0.34 & 0.39 & 0.35 & 0.41 & 0.39 & 0.41 & 0.38 & 0.43 & 0.42\\
& Crossformer & 0.32 & 0.23 & 0.41 & 0.37 & 0.42 & 0.37 & 0.51 & 0.51 & 0.53 & 0.52 & 0.58 & 0.61\\
& PatchTST & 0.29 & 0.21 & \textbf{0.35} & 0.32 & 0.36 & 0.34 & \textbf{0.38} & 0.38 & 0.38 & 0.38 & 0.40 & 0.41\\
      \cmidrule{2-14}
      & HUTFormer & \textbf{0.28} & \textbf{0.20} & \textbf{0.35} & \textbf{0.31} & \textbf{0.35} & \textbf{0.31} & \textbf{0.38} & \textbf{0.36} & \textbf{0.36} & \textbf{0.35} & \textbf{0.38} & \textbf{0.39}\\
    \midrule
    \midrule
    \multirow{7}*{\rotatebox{90}{Weather}}
& Informer & 0.34 & 0.27 & 0.38 & 0.36 & 0.40 & 0.38 & 0.43 & 0.42 & 0.45 & 0.46 & 0.45 & 0.48\\
& Autoformer & 0.36 & 0.29 & 0.38 & 0.33 & 0.39 & 0.35 & 0.41 & 0.39 & 0.42 & 0.41 & 0.44 & 0.44\\
& FEDformer & 0.32 & 0.24 & 0.34 & 0.27 & 0.35 & 0.30 & 0.36 & 0.32 & 0.38 & 0.35 & 0.47 & 0.48\\
& Pyraformer & 0.28 & 0.23 & 0.42 & 0.45 & 0.36 & 0.34 & 0.38 & 0.38 & 0.50 & 0.59 & 0.42 & 0.44\\
& Triformer & 0.15 & 0.12 & 0.23 & 0.20 & 0.26 & 0.22 & 0.28 & 0.24 & 0.32 & 0.29 & 0.34 & 0.33\\
& Crossformer & 0.14 & 0.11 & 0.22 & 0.19 & 0.25 & 0.21 & 0.27 & \textbf{0.23} & 0.31 & 0.27 & 0.32 & 0.31\\
& PatchTST & 0.14 & 0.11 & 0.22 & 0.18 & 0.25 & 0.21 & 0.27 & \textbf{0.23} & 0.31 & 0.28 & 0.33 & 0.32\\
      \cmidrule{2-14}
      & HUTFormer & \textbf{0.12} & \textbf{0.10} & \textbf{0.20} & \textbf{0.16} & \textbf{0.24} & \textbf{0.20} & \textbf{0.26} & \textbf{0.23} & \textbf{0.29} & \textbf{0.27} & \textbf{0.31} & \textbf{0.30}\\
    \midrule

    \toprule
    \end{tabular}
    }
  \end{table*}

The ability of HUTFormer to generate and utilize multi-scale features should also be effective in many non-traffic data, since the multi-scale features widely exists in many domains. In order to verify the generalization of HUTFormer, in this part, we compare HUTFormer with more latest Transformer-based long time series forecasting models~\citep{2023Crossformer, Triformer} based on three commonly used long-sequence prediction datasets, ETTh1, ETTm1, and Weather. The details of Crossformer~\citep{2023Crossformer} and Triformer~\citep{Triformer} as well as the three datasets are neglected for simplicity. Interest readers can refer to their papers~\citep{2023Crossformer, Triformer}.
We use the same setting as the other datasets in our paper.
As shown in the Table \ref{tab:more_datasets}, HUTFormer still outperforms these models on these datasets, which further verifies the effectiveness and generalization of HUTFormer.

\subsection{Ablation study}
\label{section:ablation}

In this subsection, we conduct more experiments to evaluate the impact of some important components and strategies.
Specifically, we evaluate from three aspects, including the effectiveness of the hierarchical U-net structure, the input embedding strategy, and the two-stage training strategy.
Due to space limitations, we only present the results on METR-LA datasets in Table \ref{tab:ablation}.

The hierarchical U-net structure is designed to generate and exploit multi-scale features.
Specifically, the encoder combines window self-attention and segment merging to generate multi-scale features, while the decoder primarily utilizes extracted features based on cross-scale attention.
Therefore, to evaluate their effectiveness, we set up three variants.
{\color{black}
First, we replace the decoder with a simple concatenation, named HUTFormer concat.
The concatenation of features from different scales naturally preserves all information.
Second, we set HUTFormer w/o decoder to remove the decoder and use the intermediate prediction as the final prediction.
The above two variants are used to demonstrate that exploiting multi-scale features is a non-trivial challenge and our hierarchical decoder is effective.}
Third, we set HUTFormer w/o hierarchy to further remove segment merging and replace the window Transformer layer with a standard Transformer layer, to evaluate the effectiveness of hierarchical multi-scale representations.
As shown in Table \ref{tab:ablation}, 
HUTFormer significantly outperforms HUTFormer concat and HUTFormer w/o decoder, which shows that it is not an easy task to utilize the multi-scale features, and validates the effectiveness of our decoder.
In addition, HUTFormer w/o hierarchy shows that hierarchical multi-scale features are crucial for accurate long-term traffic forecasting.
The above results show that generating and utilizing hierarchical multi-scale features is important, and the designed hierarchical U-net structure is effective.

\begin{table*}[t]
    \centering
    \renewcommand\arraystretch{1}
    \caption{
    Ablation study on the METR-LA dataset.
    }
    \label{tab:ablation}
    \scalebox{0.82}{
    \begin{tabular}{ccc|cc|cc|cc|cc|cc}
      \toprule
      \midrule
      \multirow{2}*{Variant} & 
      \multicolumn{2}{c}{@Horizon 12}& 
      \multicolumn{2}{c}{@Horizon 48}& 
      \multicolumn{2}{c}{@Horizon 96}& 
      \multicolumn{2}{c}{@Horizon 144}& 
      \multicolumn{2}{c}{@Horizon 192}& 
      \multicolumn{2}{c}{@Horizon 288} \\
      \cmidrule(r){2-3} \cmidrule(r){4-5} \cmidrule(r){6-7} \cmidrule(r){8-9} \cmidrule(r){10-11} \cmidrule(r){12-13}
      & MAE & MAPE (\%) & MAE & MAPE (\%) & MAE & MAPE (\%) & MAE & MAPE (\%) & MAE & MAPE (\%) & MAE & MAPE(\%)\\
      \midrule
      HUTFormer & \textbf{3.59} & \textbf{10.93} & \textbf{3.77} & \textbf{11.88} & \textbf{3.79} & \textbf{11.86} & \textbf{3.80} & \textbf{12.08} & \textbf{3.82} & \textbf{12.18} & \textbf{3.84} & \textbf{12.28}\\
      \cmidrule{1-13}
      concat & 3.86 & 12.16 & 3.98 & 13.23 & 4.01 & 13.36 & 4.01 & 13.36 & 4.05 & 13.41 & 4.08 & 13.65\\
      \makecell[c]{w/o  decoder} & 3.80 & 11.94 & 3.85 & 12.33 & 3.90 & 12.64 & 3.88 & 12.91 & 3.96 & 12.91 & 3.97 & 12.93\\
      \makecell[c]{w/o  hierarchy} & 3.90 & 12.56 & 3.97 & 12.85 & 3.96 & 12.86 & 3.98 & 12.88 & 3.98 & 12.92 & 4.12 & 13.48\\
      \cmidrule{1-13}
      w/o ST-PE & 4.11 & 12.68 & 4.78 & 15.80 & 4.90 & 16.44 & 5.00 & 16.81 & 5.13 & 17.47 & 5.25 & 17.56\\
      GCN & 3.79 & 11.87 & 4.23 & 14.14 & 4.28 & 14.32 & 4.30 & 14.21 & 4.32 & 14.28 & 4.35 & 14.40\\
      w/o SE & 3.76 & 11.83 & 3.86 & 12.39 & 3.85 & 12.35 & 3.91 & 12.73 & 3.92 & 12.75 & 4.03 & 13.12\\
      \cmidrule{1-13}
      end2end & 3.72 & 11.60 & 3.95 & 12.59 & 3.97 & 12.83 & 3.95 & 12.58 & 3.95 & 12.58 & 4.00 & 12.73\\
      w/o fix & 3.64 & 11.28 & 3.85 & 12.11 & 3.88 & 12.49 & 3.90 & 12.40 & 3.93 & 12.57 & 3.91 & 12.57\\
      \cmidrule{1-13}
    \toprule
    \end{tabular}
    }
  \end{table*}

The input embedding strategy aims to address the complexity issue from both spatial and temporal dimensions.
Specifically, it consists of a Segment Embedding~(SE) and a Spatial-Temporal Positional Encoding~(ST-PE).
To verify their effectiveness, we set up three variants.
First, we set HUTFormer w/o ST-PE, which replaces the ST-PE with standard learnable positional encoding.
Second, we set HUTFormer GCN, which replaces the spatial embeddings in ST-PE with graph convolution~\citep{2019GWNet}.
Third, we remove the segment embedding to get HUTFormer w/o SE.
As shown in Table \ref{tab:ablation}, 
without ST-PE, 
the performance of HUTFormer decreases significantly.
This is because modeling the correlations between time series is the basis of traffic forecasting.
In addition, we can see that the ST-PE strategy is significantly better than performing graph convolution, indicating the superiority of ST-PE.
Moreover, removing segment embedding not only leads to a significant decrease in performance but also increases the complexity due to the increased sequence length.
{\color{black}
These results indicate the effectiveness of the spatial-temporal positional encoding and segment merging.}


Finally, we evaluate the two-stage training strategy of HUTFormer.
To this end, we set two variants.
First, we set HUTFormer end2end, which trains the HUTFormer in an end-to-end strategy.
Second, we set HUTFormer w/o fix, which does not fix the parameter of the encoder when training the decoder.
The results in Table \ref{tab:ablation} show that either the end-to-end strategy or the strategy without fixing the encoder leads to insufficient optimization and significant performance degradation.
In addition, both strategies require more memory.
In contrast, our two-stage strategy achieves the best performance and efficiency simultaneously.

\subsection{Hyper-parameter and convergence study}
\label{sec:hyper}

\begin{figure}[h]
  \centering
  \includegraphics[width=0.9\linewidth]{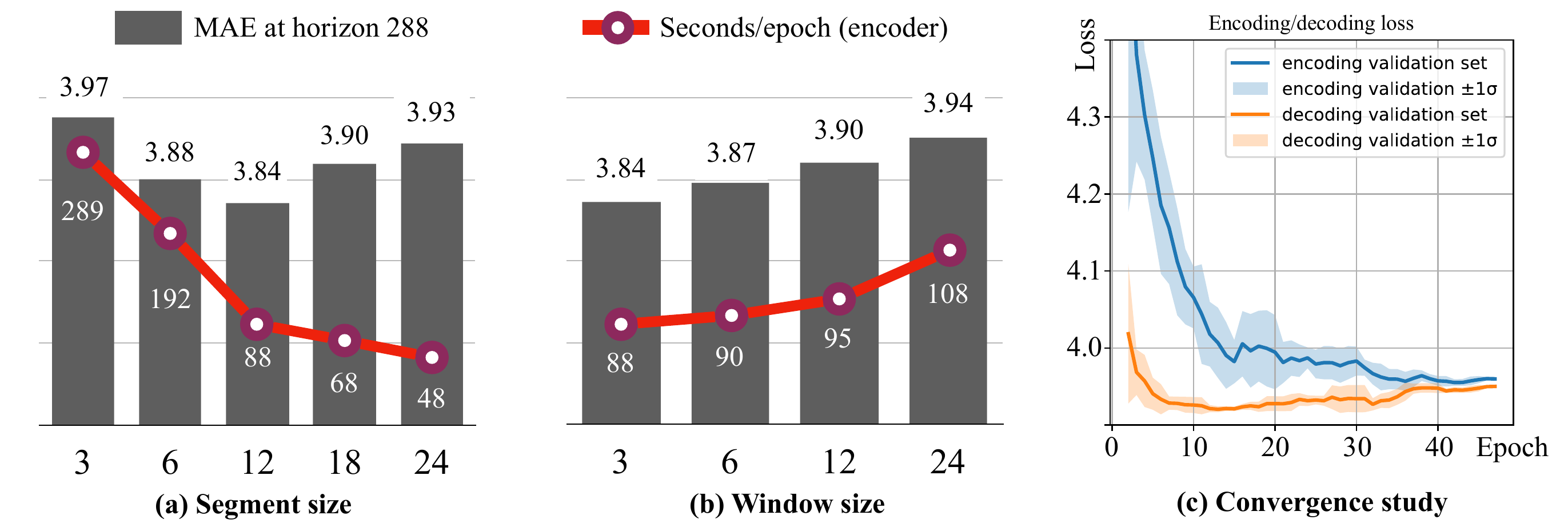}
  \caption{
  Hyper-parameter and converagence study.}
  \label{figure:vis_hyper}
\end{figure}

In this subsection, we first conduct experiments to study the impact of two key hyper-parameters: segment size and window size.
We conduct experiments on the METR-LA dataset and report the MAE at horizon 288.
Moreover, we report the training speed of the encoder, since these hyper-parameters mainly affect the encoder.
As shown in Fig. \ref{figure:vis_hyper}a, the segment size $L=12$ achieves the best performance. Smaller segments cannot provide robust semantics, while larger segments ignore more local details.
In addition, we can see that as the segment size increases, the encoder runs faster~(s/epoch).
Kindly note that changing the segment size may change the depth of the HUTFormer to ensure that the receptive field covers the entire sequence.
The impact of the window size is shown in Fig. \ref{figure:vis_hyper}b, where larger window sizes perform worse.
This is because the ability to extract multi-scale features is weakened as the window size increases.
Moreover, the efficiency of HUTFormer will also decrease~\citep{2021SwinTransformer} on larger window sizes.

{\color{black}
Additionally, we conduct a convergence analysis experiment. Fig. \ref{figure:vis_hyper}c illustrates the validation set loss during the two-phase process, showing the convergence behavior of the encoding and decoding stages. Combined with Table~\ref{tab:ablation} and Fig.~\ref{figure:vis}, it can be observed that the predictions during the encoding phase already achieve good accuracy. The decoding phase, by introducing multi-scale information, further improves the prediction results, especially in the details (particularly in periods of traffic congestion), leading to a further reduction in loss.
}

\subsection{Visualization}

\subsubsection{Spatial-temporal positional encoding}
To further understand the HUTFormer in modeling the correlations between multiple time series in traffic data, we analyze the spatial-temporal positional encoding layer.
Modeling correlations between multiple time series have been widely discussed in multivariate time series forecasting~\citep{2019GWNet, 2020MTGNN,2022STEP}.
Previous works usually utilize Graph Convolution Networks~(GCN), which conduct message passing in a pre-defined graph.
GCN is a powerful model, but it has high complexity of $\mathcal{O}(N^2)$.
Very recent works, STID~\citep{2022STID} and ST-Norm~\citep{2021STNorm}, identify that graph convolution in multivariate time series forecasting is essentially used for addressing the indistinguishability of samples on the spatial dimension.
Based on such an observation, STID proposes a simple but effective baseline of attaching spatial and temporal identities, achieving a similar performance of GCN but high efficiency.
The Spatial-Temporal Positional Encoding~(ST-PE) is designed based on such an idea~\citep{2022STID}.

The ST-PE contains three learnable positional embeddings, $\mathbf{E}\in\mathbb{R}^{N\times d}$, 
$\mathbf{T}^{TiD}\in\mathbb{R}^{N_D\times d}$, and $\mathbf{T}^{DiW}\in\mathbb{R}^{N_W\times d}$, 
where $N$ is the number of time series, $N_D$ is the number of time slots of a day~(determined by the sensor's sampling frequency), and $N_W=7$ is the number of days in a week. 
We utilize t-SNE~\citep{2008T-SNE}. to visualize these three embedding matrices.
Kindly note that $\mathbf{T}^{DiW}$ only have $7$ embeddings, which is significantly less than the hidden dimension $32$, making it hard to get correct visualizations.
Therefore, we additionally train a HUTFormer with the embedding size of  $\mathbf{T}^{DiW}$ to 2 to get a more accurate visualization.

\begin{figure*}
  \centering
  \includegraphics[width=1\linewidth]{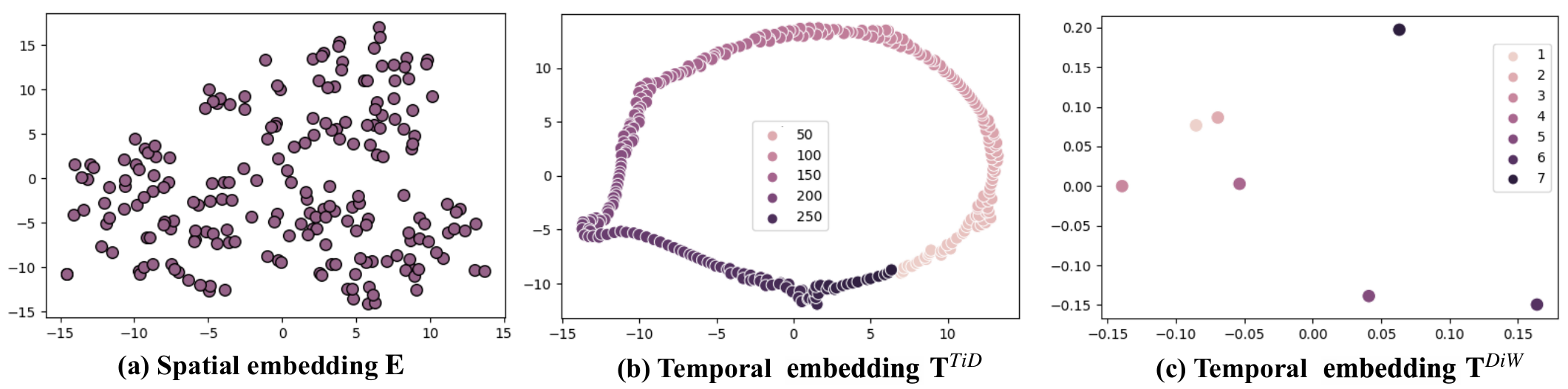}
  \caption{
  Visualization of the spatial and temporal embeddings.}
  \label{figure:vis_stid}
\end{figure*}

The results are shown in Fig. \ref{figure:vis_stid}.
First, as shown in Fig. \ref{figure:vis_stid}a, the spatial embeddings are likely to cluster.
For example, traffic conditions observed by sensors that are connected or have similar geographical functionality are more likely to be similar.
However, it is not as apparent as in the results in STID~\citep{2022STID}.
We conjecture this is because the impact of the indistinguishability of the samples becomes weaker as the length of the historical data increases.
Second, Fig. \ref{figure:vis_stid}b shows the embeddings of 288 time slots, where the daily periodicity is very obvious.
Third, Fig. \ref{figure:vis_stid}c visualizes the embeddings of each day in a week, where weekdays are closer and weekends' are different.

\begin{figure}[htbp]
  \centering
  \includegraphics[width=1\linewidth]{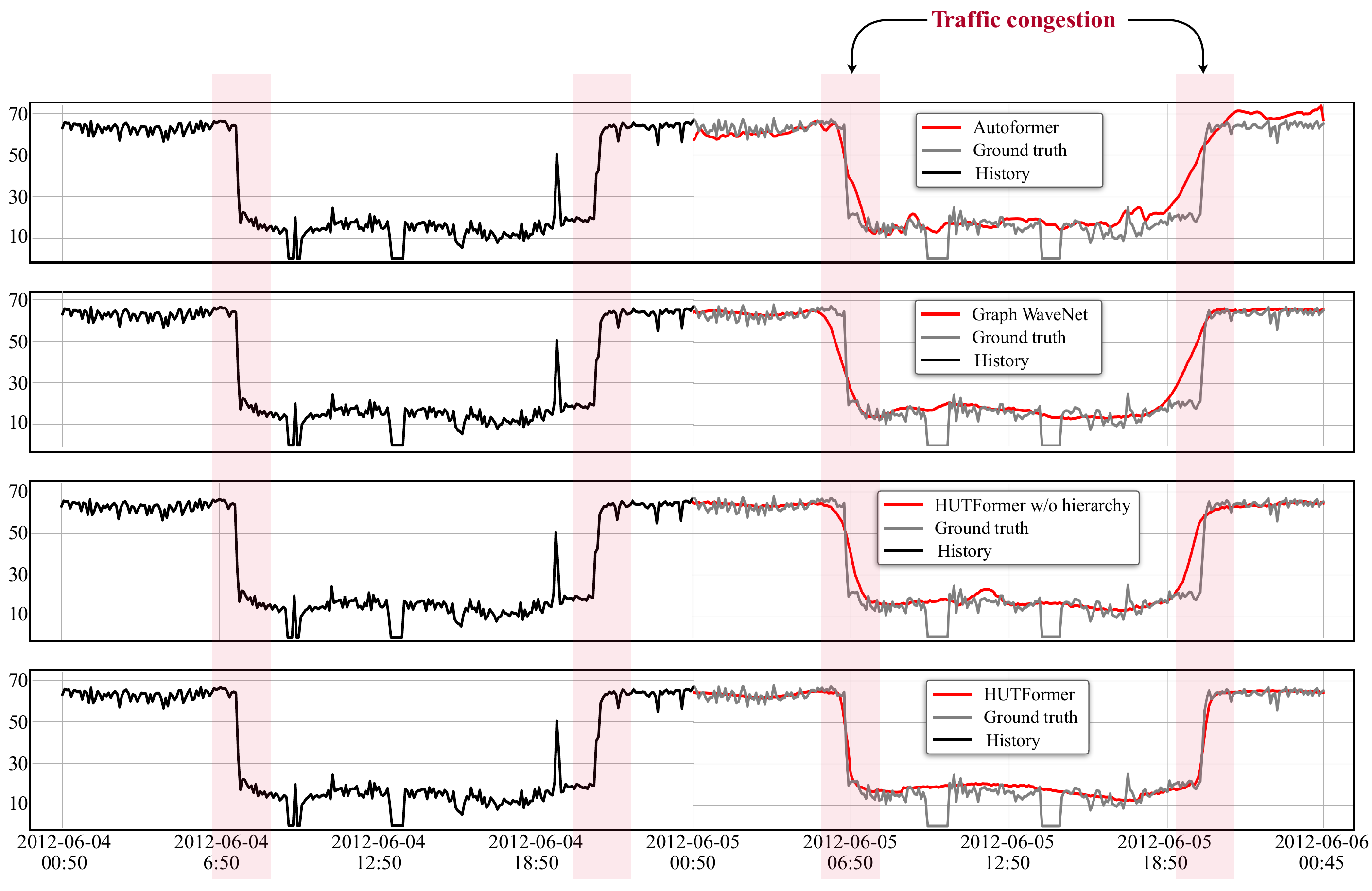}
  \caption{
  Visualization of prediction results of HUTFormer and some baseline models.}
  \label{figure:vis}
\end{figure}

\subsubsection{Prediction visualization}
{\color{black}
In order to further intuitively evaluate HUTFormer, in this subsection, we visualize the prediction of HUTFormer and other baselines on the METR-LA dataset.
Specifically, we select sensor 12 and displayed its data from June 05th, 2012 to June 06th, 2012 (located in the test dataset). 

Fig. \ref{figure:vis} shows two consecutive days from the METR-LA dataset: the first day is used as historical input, while the second day is the prediction target. This design serves two purposes. First, the strong similarity between the two days illustrates the dataset’s pronounced periodicity. Second, the series captures sharp, localized fluctuations during the morning and evening rush hours, revealing the onset and dissipation of traffic congestion.

Capturing global patterns is essential for modeling overall cyclical trends, whereas capturing local patterns is crucial for accurately forecasting fine-grained changes such as rapid, short-term spikes. As discussed previously, models like Autoformer~\citep{2021AutoFormer}, Graph WaveNet~\citep{2019GWNet}, and HUTFormer w/o hierarchy focus primarily on global dynamics and largely overlook local feature extraction. Consequently, although they reproduce broad periodic behavior reasonably well, their accuracy falls during intervals of rapid change—specifically, the congestion periods marked by the red background in the figure. By contrast, HUTFormer effectively models multi-scale features and thus maintains high accuracy even in these volatile segments. This result underscores the importance of integrating features at multiple scales when modeling complex, periodic traffic data.
}

\subsection{Limitations}
{\color{black}
Although HUTFormer demonstrates strong performance, it still has several issues that need to be addressed in future work.
First, the most significant issue with HUTFormer is its lack of generalization to new sensors. Graph-based spatial dependency modeling methods~\citep{2017GCN, 2018DCRNN, 2019GWNet} are inherently inductive~\citep{GraphSAGE}, meaning they can make predictions on graphs with changing nodes and relationships. However, HUTFormer relies on spatial positional encoding to capture spatial dependencies. For newly introduced nodes, the positional encoding requires training, which means HUTFormer cannot naturally perform inductive reasoning.
Second, the training of HUTFormer is more complex than that of end-to-end methods. Although the two-phase training strategy is effective, it objectively makes the model's training process less convenient compared to end-to-end models. Therefore, exploring methods for end-to-end generation and utilization of multi-scale information is a promising avenue for future research.
In addition, the reliability of longer-term forecasting is crucial. The longer the forecast period, the greater the uncertainty. Only by systematically quantifying these uncertainties and incorporating known external events into the model can more accurate and practically valuable results be achieved.
}
\section{Conclusions}

In this study, we make the first attempt to explore the long-term traffic forecasting problem.
To this end, we reveal its unique challenges in exploiting the multi-scale representations of traffic data, and propose a novel Hierarchical U-net TransFormer~(HUTFormer) to efficiently and effectively address them.
The HUTFormer mainly consists of a hierarchical encoder and decoder.
On the one hand, the hierarchical encoder generates multi-scale representations based on the window self-attention mechanism and segment merging. 
On the other hand, the hierarchical decoder effectively utilizes the extracted multi-scale features based on the cross-scale attention mechanism. 
In addition, HUTFormer adopts segment embedding and spatial-temporal positional encoding as the input embedding strategy to address the complexity issue. 
Extensive experiments on four commonly used traffic datasets show that the proposed HUTFormer significantly outperforms state-of-the-art traffic forecasting and long-sequence time series forecasting baselines.

\section*{Acknowledgements}

This work is supported by NSFC No.62502505, No. 62372430, and No.62502499, the Youth Innovation Promotion Association CAS No.2023112, the Postdoctoral Fellowship Program of CPSF under Grant Number GZC20251078 and Number GZC20241758 and the China Postdoctoral Science Foundation No.2025M771542.

\section*{References}
\nocite{*}
\bibliographystyle{elsarticle-harv}
\bibliography{test}

@inproceedings{2025BLAST,
  title={BLAST: Balanced Sampling Time Series Corpus for Universal Forecasting Models},
  author={Shao, Zezhi and Li, Yujie and Wang, Fei and Yu, Chengqing and Fu, Yisong and Qian, Tangwen and Xu, Bin and Diao, Boyu and Xu, Yongjun and Cheng, Xueqi},
  booktitle={Proceedings of the 31st ACM SIGKDD Conference on Knowledge Discovery and Data Mining V. 2},
  pages={2502--2513},
  year={2025}
}

@inproceedings{GraphSAGE,
  author       = {William L. Hamilton and
                  Zhitao Ying and
                  Jure Leskovec},
  title        = {Inductive Representation Learning on Large Graphs},
  booktitle    = {Advances in Neural Information Processing Systems},
  pages        = {1024--1034},
  year         = {2017},
}

@article{liu2021deeptsp,
  title={DeepTSP: Deep traffic state prediction model based on large-scale empirical data},
  author={Liu, Yang and Lyu, Cheng and Zhang, Yuan and Liu, Zhiyuan and Yu, Wenwu and Qu, Xiaobo},
  journal={Commun Transp Res},
  volume={1},
  pages={100012},
  year={2021},
}

@article{IF,
  author       = {Ying Li and
                  Fan Bai and
                  Cheng Lyu and
                  Xiaobo Qu and
                  Yang Liu},
  title        = {A systematic review of generative adversarial networks for traffic
                  state prediction: Overview, taxonomy, and future prospects},
  journal      = {Inf. Fusion},
  volume       = {117},
  pages        = {102915},
  year         = {2025},
  url          = {https://doi.org/10.1016/j.inffus.2024.102915},
  doi          = {10.1016/J.INFFUS.2024.102915},
  bibsource    = {dblp computer science bibliography, https://dblp.org}
}

@article{BasicTS,
  author       = {Zezhi Shao and
                  Fei Wang and
                  Yongjun Xu and
                  Wei Wei and
                  Chengqing Yu and
                  Zhao Zhang and
                  Di Yao and
                  Tao Sun and
                  Guangyin Jin and
                  Xin Cao and
                  Gao Cong and
                  Christian S. Jensen and
                  Xueqi Cheng},
  title        = {Exploring Progress in Multivariate Time Series Forecasting: Comprehensive
                  Benchmarking and Heterogeneity Analysis},
  journal      = {IEEE Trans Knowl Data Eng},
  volume       = {37},
  number       = {1},
  pages        = {291--305},
  year         = {2025},
}

@article{bogaerts2020graph,
  title={A graph CNN-LSTM neural network for short and long-term traffic forecasting based on trajectory data},
  author={Bogaerts, Toon and Masegosa, Antonio D and Angarita-Zapata, Juan S and Onieva, Enrique and Hellinckx, Peter},
  journal={Transp Res Part C Emerg Technol},
  volume={112},
  pages={62--77},
  year={2020},
}

@article{2020AppliedIntelligence,
  title={A recurrent neural network for urban long-term traffic flow forecasting},
  author={Belhadi, Asma and Djenouri, Youcef and Djenouri, Djamel and Lin, Jerry Chun-Wei},
  journal={Appl Intell},
  volume={50},
  pages={3252--3265},
  year={2020},
}

@article{2021TITS,
  title={Long-term traffic prediction based on lstm encoder-decoder architecture},
  author={Wang, Zhumei and Su, Xing and Ding, Zhiming},
  journal={IEEE Trans Intell Transp Syst},
  volume={22},
  number={10},
  pages={6561--6571},
  year={2020},
}

@article{COMMTR1,
  title={Towards explainable traffic flow prediction with large language models},
  author={Guo, Xusen and Zhang, Qiming and Jiang, Junyue and Peng, Mingxing and Zhu, Meixin and Yang, Hao Frank},
  journal={Commun Transp Res},
  volume={4},
  pages={100150},
  year={2024},
}

@article{COMMTR2,
  title={AGNP: Network-wide short-term probabilistic traffic speed prediction and imputation},
  author={Xu, Meng and Di, Yining and Ding, Hongxing and Zhu, Zheng and Chen, Xiqun and Yang, Hai},
  journal={Commun Transp Res},
  volume={3},
  pages={100099},
  year={2023},
}

@article{TRE3,
  title={A non-additive path-based reward credit scheme for traffic congestion management},
  author={Luan, Mingye and Waller, S Travis and Rey, David},
  journal={Transp Res Part E Logist Transp Rev},
  volume={179},
  pages={103291},
  year={2023},
}

@article{TRE2,
  title={Real-time demand forecasting for an urban delivery platform},
  author={Hess, Alexander and Spinler, Stefan and Winkenbach, Matthias},
  journal={Transp Res Part E Logist Transp Rev},
  volume={145},
  pages={102147},
  year={2021},
}

@article{TRE1,
  title={TS-STNN: Spatial-temporal neural network based on tree structure for traffic flow prediction},
  author={Lv, Yang and Lv, Zhiqiang and Cheng, Zesheng and Zhu, Zhanqi and Rashidi, Taha Hossein},
  journal={Transp Res Part E Logist Transp Rev},
  volume={177},
  pages={103251},
  year={2023},
}

@inproceedings{DFGCN,
  title={Dynamic Frequency Domain Graph Convolutional Network for Traffic Forecasting},
  author={Li, Yujie and Shao, Zezhi and Xu, Yongjun and Qiu, Qiang and Cao, Zhaogang and Wang, Fei},
  booktitle={ICASSP 2024-2024 IEEE International Conference on Acoustics, Speech and Signal Processing (ICASSP)},
  pages={5245--5249},
  year={2024},
}

@inproceedings{CANet,
  title={Clustering-property Matters: A Cluster-aware Network for Large Scale Multivariate Time Series Forecasting},
  author={Wang, Yuan and Shao, Zezhi and Sun, Tao and Yu, Chengqing and Xu, Yongjun and Wang, Fei},
  booktitle={Proceedings of the 32nd ACM International Conference on Information and Knowledge Management},
  pages={4340--4344},
  year={2023}
}

@inproceedings{GinAR,
  author       = {Chengqing Yu and
                  Fei Wang and
                  Zezhi Shao and
                  Tangwen Qian and
                  Zhao Zhang and
                  Wei Wei and
                  Yongjun Xu},
  title        = {GinAR: An End-To-End Multivariate Time Series Forecasting Model Suitable for Variable Missing},
  booktitle    = {Proceedings of the 30th {ACM} {SIGKDD} Conference on Knowledge Discovery and Data Mining},
  pages        = {3989--4000},
  year         = {2024},
}

@inproceedings{DSformer,
  author       = {Chengqing Yu and
                  Fei Wang and
                  Zezhi Shao and
                  Tao Sun and
                  Lin Wu and
                  Yongjun Xu},
  title        = {DSformer: {A} Double Sampling Transformer for Multivariate Time Series
                  Long-term Prediction},
  booktitle    = {Proceedings of the 32nd {ACM} International Conference on Information
                  and Knowledge Management},
  pages        = {3062--3072},
  year         = {2023},
}

@article{tiv1,
  author       = {Liang Chu and
                  Zhuoran Hou and
                  Jingjing Jiang and
                  Jun Yang and
                  Yuanjian Zhang},
  title        = {Spatial-Temporal Feature Extraction and Evaluation Network for Citywide
                  Traffic Condition Prediction},
  journal      = {IEEE Trans Intell Veh},
  volume       = {9},
  number       = {9},
  pages        = {5377--5391},
  year         = {2024},
}

@article{tiv2,
  title={Semi-persistent resource allocation based on traffic prediction for vehicular communications},
  author={Chu, Ping and Zhang, J Andrew and Wang, Xiaoxiang and Fang, Gengfa and Wang, Dongyu},
  journal={IEEE Trans Intell Veh},
  volume={5},
  number={2},
  pages={345--355},
  year={2019},
}

@inproceedings{2023PatchTST,
  author       = {Yuqi Nie and
                  Nam H. Nguyen and
                  Phanwadee Sinthong and
                  Jayant Kalagnanam},
  title        = {A Time Series is Worth 64 Words: Long-term Forecasting with Transformers},
  booktitle    = {International Conference on Learning Representations},
  year         = {2023},
  pages        = {1-24}
}

@inproceedings{Scaleformer,
  author       = {Mohammad Amin Shabani and
                  Amir H. Abdi and
                  Lili Meng and
                  Tristan Sylvain},
  title        = {Scaleformer: Iterative Multi-scale Refining Transformers for Time Series Forecasting},
  booktitle    = {International Conference on Learning Representations},
  year         = {2023},
  pages        = {1-23}
}

@inproceedings{Triformer,
  author       = {Razvan{-}Gabriel Cirstea and
                  Chenjuan Guo and
                  Bin Yang and
                  Tung Kieu and
                  Xuanyi Dong and
                  Shirui Pan},
  title        = {Triformer: Triangular, Variable-Specific Attentions for Long Sequence
                  Multivariate Time Series Forecasting},
  booktitle    = {Proceedings of the Thirty-First International Joint Conference on Artificial Intelligence},
  pages        = {1994--2001},
  year         = {2022},
}

@article{shuman2013emerging,
  title={The emerging field of signal processing on graphs: Extending high-dimensional data analysis to networks and other irregular domains},
  author={Shuman, David I and Narang, Sunil K and Frossard, Pascal and Ortega, Antonio and Vandergheynst, Pierre},
  journal={IEEE Signal Process Mag},
  volume={30},
  number={3},
  pages={83--98},
  year={2013},
}

@article{innovation1,
title = {Spatial-temporal large models: A super hub linking multiple scientific areas with artificial intelligence},
journal = {Innovation},
volume = {6},
number = {2},
pages = {100763},
year = {2025},
author = {Zezhi Shao and Tangwen Qian and Tao Sun and Fei Wang and Yongjun Xu}
}

@article{innovation2,
  title={AI-enhanced spatial-temporal data-mining technology: New chance for next-generation urban computing},
  author={Wang, Fei and Yao, Di and Li, Yong and Sun, Tao and Zhang, Zhao},
  journal={Innovation},
  volume={4},
  number={2},
  year={2023},
  pages = {100405},
}

@article{liu2019deeppf,
  title={DeepPF: A deep learning based architecture for metro passenger flow prediction},
  author={Liu, Yang and Liu, Zhiyuan and Jia, Ruo},
  journal={Transp Res Part C Emerg Technol},
  volume={101},
  pages={18--34},
  year={2019},
}

@article{innovation3,
  title={Foundation models and intelligent decision-making: Progress, challenges, and perspectives},
  author={Huang, Jincai and Xu, Yongjun and Wang, Qi and Wang, Qi Cheems and Liang, Xingxing and Wang, Fei and Zhang, Zhao and Wei, Wei and Zhang, Boxuan and Huang, Libo and others},
  pages={100948},
  volume={6},
  issue={6},
  journal={Innovation},
  year={2025},
}

@inproceedings{2023Crossformer,
  author       = {Yunhao Zhang and
                  Junchi Yan},
  title        = {Crossformer: Transformer Utilizing Cross-Dimension Dependency for Multivariate Time Series Forecasting},
  booktitle    = {International Conference on Learning Representations},
  year         = {2023},
  pages        = {1-21}
}

@inproceedings{2017FPN,
  title={Feature pyramid networks for object detection},
  author={Lin, Tsung-Yi and Doll{\'a}r, Piotr and Girshick, Ross and He, Kaiming and Hariharan, Bharath and Belongie, Serge},
  booktitle={Proceedings of the IEEE conference on computer vision and pattern recognition},
  pages={2117--2125},
  year={2017}
}

@inproceedings{2015UNet,
  title={U-net: Convolutional networks for biomedical image segmentation},
  author={Ronneberger, Olaf and Fischer, Philipp and Brox, Thomas},
  booktitle={International Conference on Medical image computing and computer-assisted intervention},
  pages={234--241},
  year={2015},
}

@inproceedings{2021SwinUNet,
  author       = {Hu Cao and
                  Yueyue Wang and
                  Joy Chen and
                  Dongsheng Jiang and
                  Xiaopeng Zhang and
                  Qi Tian and
                  Manning Wang},
  title        = {Swin-Unet: Unet-Like Pure Transformer for Medical Image Segmentation},
  booktitle    = {Computer Vision - {ECCV} 2022 Workshops},
  volume       = {13803},
  pages        = {205--218},
  year         = {2022},
}

@inproceedings{2016ResNet,
  title={Deep residual learning for image recognition},
  author={He, Kaiming and Zhang, Xiangyu and Ren, Shaoqing and Sun, Jian},
  booktitle={Proceedings of the IEEE conference on computer vision and pattern recognition},
  pages={770--778},
  year={2016}
}

@inproceedings{2021ViT,
  author       = {Alexey Dosovitskiy and
                  Lucas Beyer and
                  Alexander Kolesnikov and
                  Dirk Weissenborn and
                  Xiaohua Zhai and
                  Thomas Unterthiner and
                  Mostafa Dehghani and
                  Matthias Minderer and
                  Georg Heigold and
                  Sylvain Gelly and
                  Jakob Uszkoreit and
                  Neil Houlsby},
  title        = {An Image is Worth 16x16 Words: Transformers for Image Recognition at Scale},
  booktitle    = {International Conference on Learning Representations},
  year         = {2021},
  pages        = {1-24}
}

@inproceedings{2021SwinTransformer,
  title={Swin transformer: Hierarchical vision transformer using shifted windows},
  author={Liu, Ze and Lin, Yutong and Cao, Yue and Hu, Han and Wei, Yixuan and Zhang, Zheng and Lin, Stephen and Guo, Baining},
  booktitle={Proceedings of the IEEE/CVF international conference on computer vision},
  pages={10012--10022},
  year={2021}
}

@inproceedings{2019STMetaNet,
  title={Urban traffic prediction from spatio-temporal data using deep meta learning},
  author={Pan, Zheyi and Liang, Yuxuan and Wang, Weifeng and Yu, Yong and Zheng, Yu and Zhang, Junbo},
  booktitle={Proceedings of the 25th ACM SIGKDD international conference on knowledge discovery \& data mining},
  pages={1720--1730},
  year={2019}
}

@inproceedings{2022STEP,
  title={Pre-training enhanced spatial-temporal graph neural network for multivariate time series forecasting},
  author={Shao, Zezhi and Zhang, Zhao and Wang, Fei and Xu, Yongjun},
  booktitle={Proceedings of the 28th ACM SIGKDD conference on knowledge discovery and data mining},
  pages={1567--1577},
  year={2022}
}

@inproceedings{2022STID,
  title={Spatial-temporal identity: A simple yet effective baseline for multivariate time series forecasting},
  author={Shao, Zezhi and Zhang, Zhao and Wang, Fei and Wei, Wei and Xu, Yongjun},
  booktitle={Proceedings of the 31st ACM international conference on information \& knowledge management},
  pages={4454--4458},
  year={2022}
}

@inproceedings{2019GWNet,
  author       = {Zonghan Wu and
                  Shirui Pan and
                  Guodong Long and
                  Jing Jiang and
                  Chengqi Zhang},
  title        = {Graph WaveNet for Deep Spatial-Temporal Graph Modeling},
  booktitle    = {Proceedings of the Twenty-Eighth International Joint Conference on Artificial Intelligence},
  pages        = {1907--1913},
  year         = {2019}
}

@inproceedings{2018DCRNN,
  author    = {Yaguang Li and
               Rose Yu and
               Cyrus Shahabi and
               Yan Liu},
  title     = {Diffusion Convolutional Recurrent Neural Network: Data-Driven Traffic Forecasting},
  booktitle = {International Conference on Learning Representations},
  year      = {2018},
  pages     = {1-16}
}

@article{2022D2STGNN,
  title={Decoupled dynamic spatial-temporal graph neural network for traffic forecasting},
  author={Shao, Zezhi and Zhang, Zhao and Wei, Wei and Wang, Fei and Xu, Yongjun and Cao, Xin and Jensen, Christian S},
  journal={Proc VLDB Endow},
  volume={15},
  number={11},
  pages={2733--2746},
  year={2022},
}

@inproceedings{2020AGCRN,
  author       = {Lei Bai and
                  Lina Yao and
                  Can Li and
                  Xianzhi Wang and
                  Can Wang},
  title        = {Adaptive Graph Convolutional Recurrent Network for Traffic Forecasting},
  booktitle    = {Advances in Neural Information Processing Systems},
  year         = {2020},
  pages={17804--17815}
}

@inproceedings{2021STNorm,
  title={St-norm: Spatial and temporal normalization for multi-variate time series forecasting},
  author={Deng, Jinliang and Chen, Xiusi and Jiang, Renhe and Song, Xuan and Tsang, Ivor W},
  booktitle={Proceedings of the 27th ACM SIGKDD conference on knowledge discovery \& data mining},
  pages={269--278},
  year={2021}
}

@article{kumar2015short,
  title={Short-term traffic flow prediction using seasonal ARIMA model with limited input data},
  author={Kumar, S Vasantha and Vanajakshi, Lelitha},
  journal={Eur Transp Res Rev},
  volume={7},
  number={3},
  pages={1--9},
  year={2015},
}

@inproceedings{2020StemGNN,
  title={Spectral temporal graph neural network for multivariate time-series forecasting},
  author={Cao, Defu and Wang, Yujing and Duan, Juanyong and Zhang, Ce and Zhu, Xia and Huang, Congrui and Tong, Yunhai and Xu, Bixiong and Bai, Jing and Tong, Jie and others},
  booktitle={Advances in neural information processing systems},
  pages={17766--17778},
  year={2020}
}

@inproceedings{2018STGCN,
  author    = {Bing Yu and
               Haoteng Yin and
               Zhanxing Zhu},
  title     = {Spatio-Temporal Graph Convolutional Networks: {A} Deep Learning Framework
               for Traffic Forecasting},
  booktitle = {IJCAI},
  pages     = {3634--3640},
  year      = {2018}
}

@inproceedings{2020STGNN,
  title={Traffic flow prediction via spatial temporal graph neural network},
  author={Wang, Xiaoyang and Ma, Yao and Wang, Yiqi and Jin, Wei and Wang, Xin and Tang, Jiliang and Jia, Caiyan and Yu, Jian},
  booktitle={Proceedings of the web conference 2020},
  pages={1082--1092},
  year={2020}
}

@article{2018T-GCN,
  author       = {Ling Zhao and
                  Yujiao Song and
                  Chao Zhang and
                  Yu Liu and
                  Pu Wang and
                  Tao Lin and
                  Min Deng and
                  Haifeng Li},
  title        = {{T-GCN:} {A} Temporal Graph Convolutional Network for Traffic Prediction},
  journal      = {IEEE Trans Intell Transp Syst},
  volume       = {21},
  number       = {9},
  pages        = {3848--3858},
  year         = {2020},
}

@inproceedings{2021DMSTGCN,
  title={Dynamic and multi-faceted spatio-temporal deep learning for traffic speed forecasting},
  author={Han, Liangzhe and Du, Bowen and Sun, Leilei and Fu, Yanjie and Lv, Yisheng and Xiong, Hui},
  booktitle={Proceedings of the 27th ACM SIGKDD conference on knowledge discovery \& data mining},
  pages={547--555},
  year={2021}
}

@inproceedings{2021GTS,
  author    = {Chao Shang and
               Jie Chen and
               Jinbo Bi},
  title     = {Discrete Graph Structure Learning for Forecasting Multiple Time Series},
  booktitle = {International Conference on Learning Representations},
  year      = {2021},
  pages     = {1-14}
}

@inproceedings{2020STGRET,
  title={ST-GRAT: A novel spatio-temporal graph attention networks for accurately forecasting dynamically changing road speed},
  author={Park, Cheonbok and Lee, Chunggi and Bahng, Hyojin and Tae, Yunwon and Jin, Seungmin and Kim, Kihwan and Ko, Sungahn and Choo, Jaegul},
  booktitle={Proceedings of the 29th ACM international conference on information \& knowledge management},
  pages={1215--1224},
  year={2020}
}

@article{2021ASTGNN,
  author       = {Shengnan Guo and
                  Youfang Lin and
                  Huaiyu Wan and
                  Xiucheng Li and
                  Gao Cong},
  title        = {Learning Dynamics and Heterogeneity of Spatial-Temporal Graph Data
                  for Traffic Forecasting},
  journal      = {IEEE Trans Knowl Data Eng},
  volume       = {34},
  number       = {11},
  pages        = {5415--5428},
  year         = {2022},
}

@inproceedings{2020GMAN,
  title={Gman: A graph multi-attention network for traffic prediction},
  author={Zheng, Chuanpan and Fan, Xiaoliang and Wang, Cheng and Qi, Jianzhong},
  booktitle={Proceedings of the AAAI conference on artificial intelligence},
  volume={34},
  number={01},
  pages={1234--1241},
  year={2020}
}

@article{2021DGCRN,
  author       = {Fuxian Li and
                  Jie Feng and
                  Huan Yan and
                  Guangyin Jin and
                  Fan Yang and
                  Funing Sun and
                  Depeng Jin and
                  Yong Li},
  title        = {Dynamic Graph Convolutional Recurrent Network for Traffic Prediction:
                  Benchmark and Solution},
  journal      = {ACM Trans Knowl Discov Data},
  volume       = {17},
  number       = {1},
  pages        = {9:1--9:21},
  year         = {2023},
}

@inproceedings{2019ASTGCN,
  author       = {Shengnan Guo and
                  Youfang Lin and
                  Ning Feng and
                  Chao Song and
                  Huaiyu Wan},
  title        = {Attention Based Spatial-Temporal Graph Convolutional Networks for
                  Traffic Flow Forecasting},
  booktitle    = {The Thirty-First Innovative Applications of Artificial Intelligence
                  Conference},
  pages        = {922--929},
  year         = {2019},
}

@inproceedings{2020MTGNN,
  title={Connecting the dots: Multivariate time series forecasting with graph neural networks},
  author={Wu, Zonghan and Pan, Shirui and Long, Guodong and Jiang, Jing and Chang, Xiaojun and Zhang, Chengqi},
  booktitle={Proceedings of the 26th ACM SIGKDD international conference on knowledge discovery \& data mining},
  pages={753--763},
  year={2020}
}

@article{vlahogianni2014short,
  title={Short-term traffic forecasting: Where we are and where we’re going},
  author={Vlahogianni, Eleni I and Karlaftis, Matthew G and Golias, John C},
  journal={Transp Res Part C Emerg Technol},
  volume={43},
  pages={3--19},
  year={2014},
}

@inproceedings{2019LogTrans,
  author       = {Shiyang Li and
                  Xiaoyong Jin and
                  Yao Xuan and
                  Xiyou Zhou and
                  Wenhu Chen and
                  Yu{-}Xiang Wang and
                  Xifeng Yan},
  title        = {Enhancing the Locality and Breaking the Memory Bottleneck of Transformer on Time Series Forecasting},
  booktitle    = {Advances in Neural Information Processing Systems},
  pages        = {5244--5254},
  year         = {2019},
}

@inproceedings{2020ReFormer,
  author    = {Nikita Kitaev and
               Lukasz Kaiser and
               Anselm Levskaya},
  title     = {Reformer: The Efficient Transformer},
  booktitle = {International Conference on Learning Representations},
  year      = {2020},
  pages     = {1-12}
}

@inproceedings{2021AutoFormer,
  author       = {Haixu Wu and
                  Jiehui Xu and
                  Jianmin Wang and
                  Mingsheng Long},
  title        = {Autoformer: Decomposition Transformers with Auto-Correlation for Long-Term
                  Series Forecasting},
  booktitle    = {Advances in Neural Information Processing Systems},
  pages        = {22419--22430},
  year         = {2021},
}

@inproceedings{2022FEDFormer,
  title={Fedformer: Frequency enhanced decomposed transformer for long-term series forecasting},
  author={Zhou, Tian and Ma, Ziqing and Wen, Qingsong and Wang, Xue and Sun, Liang and Jin, Rong},
  booktitle={International conference on machine learning},
  pages={27268--27286},
  year={2022},
}

@inproceedings{2022Pyraformer,
  author    = {Shizhan Liu and
               Hang Yu and
               Cong Liao and
               Jianguo Li and
               Weiyao Lin and
               Alex X. Liu and
               Schahram Dustdar},
  title     = {Pyraformer: Low-Complexity Pyramidal Attention for Long-Range Time Series Modeling and Forecasting},
  booktitle = {International Conference on Learning Representations},
  year      = {2022},
  pages     = {1-20}
}

@inproceedings{2021Informer,
  author       = {Haoyi Zhou and
                  Shanghang Zhang and
                  Jieqi Peng and
                  Shuai Zhang and
                  Jianxin Li and
                  Hui Xiong and
                  Wancai Zhang},
  title        = {Informer: Beyond Efficient Transformer for Long Sequence Time-Series
                  Forecasting},
  booktitle    = {Thirty-Fifth {AAAI} Conference on Artificial Intelligence},
  pages        = {11106--11115},
  year         = {2021},
}

@inproceedings{2023DLinear,
  author       = {Ailing Zeng and
                  Muxi Chen and
                  Lei Zhang and
                  Qiang Xu},
  editor       = {Brian Williams and
                  Yiling Chen and
                  Jennifer Neville},
  title        = {Are Transformers Effective for Time Series Forecasting?},
  booktitle    = {Thirty-Seventh {AAAI} Conference on Artificial Intelligence},
  pages        = {11121--11128},
  year         = {2023},
}

@inproceedings{2020HI,
  author       = {Yue Cui and
                  Jiandong Xie and
                  Kai Zheng},
  title        = {Historical Inertia: {A} Neglected but Powerful Baseline for Long Sequence
                  Time-series Forecasting},
  booktitle    = {The 30th {ACM} International Conference on Information and Knowledge Management},
  pages        = {2965--2969},
  year         = {2021},
}

@inproceedings{2017Transformer,
  author    = {Ashish Vaswani and
               Noam Shazeer and
               Niki Parmar and
               Jakob Uszkoreit and
               Llion Jones and
               Aidan N. Gomez and
               Lukasz Kaiser and
               Illia Polosukhin},
  title     = {Attention is All you Need},
  booktitle = {Advances in Neural Information Processing Systems},
  pages     = {5998--6008},
  year      = {2017}
}

@article{Innovation4,
title = {Can language models be used for real-world urban-delivery route optimization?},
journal = {Innovation},
volume = {4},
number = {6},
pages = {100520},
year = {2023},
issn = {2666-6758},
author = {Yang Liu and Fanyou Wu and Zhiyuan Liu and Kai Wang and Feiyue Wang and Xiaobo Qu},
}

@article{HetGNN,
  title={Heterogeneous graph neural network with multi-view representation learning},
  author={Shao, Zezhi and Xu, Yongjun and Wei, Wei and Wang, Fei and Zhang, Zhao and Zhu, Feida},
  journal={IEEE Trans Knowl Data Eng},
  volume={35},
  number={11},
  pages={11476--11488},
  year={2022},
}

@inproceedings{2017GCN,
  author    = {Thomas N. Kipf and
               Max Welling},
  title     = {Semi-Supervised Classification with Graph Convolutional Networks},
  booktitle = {International Conference on Learning Representations},
  year      = {2017},
  pages     = {1-14}
}

@inproceedings{2014GRU,
  author    = {Kyunghyun Cho and
               Bart van Merrienboer and
               Dzmitry Bahdanau and
               Yoshua Bengio},
  title     = {On the Properties of Neural Machine Translation: Encoder-Decoder Approaches},
  booktitle = {SSST@EMNLP 2014},
  pages     = {103--111},
  year      = {2014}
}

@article{METR-LA,
  author       = {H. V. Jagadish and
                  Johannes Gehrke and
                  Alexandros Labrinidis and
                  Yannis Papakonstantinou and
                  Jignesh M. Patel and
                  Raghu Ramakrishnan and
                  Cyrus Shahabi},
  title        = {Big data and its technical challenges},
  journal      = {Commun ACM},
  volume       = {57},
  number       = {7},
  pages        = {86--94},
  year         = {2014},
}

@article{PEMS-BAY,
  title={Freeway performance measurement system: mining loop detector data},
  author={Chen, Chao and Petty, Karl and Skabardonis, Alexander and Varaiya, Pravin and Jia, Zhanfeng},
  journal={Transp Res Rec},
  volume={1748},
  number={1},
  pages={96--102},
  year={2001},
}

@inproceedings{FC-LSTM,
  author       = {Ilya Sutskever and
                  Oriol Vinyals and
                  Quoc V. Le},
  title        = {Sequence to Sequence Learning with Neural Networks},
  booktitle    = {Advances in Neural Information Processing Systems},
  pages        = {3104--3112},
  year         = {2014}
}

@inproceedings{2016TCN,
  author    = {Fisher Yu and Vladlen Koltun},
  title     = {Multi-Scale Context Aggregation by Dilated Convolutions},
  booktitle = {International Conference on Learning Representations},
  year      = {2016},
  pages     = {1-13}
}

@inproceedings{2016GCN,
  author       = {Micha{\"{e}}l Defferrard and
                  Xavier Bresson and
                  Pierre Vandergheynst},
  title        = {Convolutional Neural Networks on Graphs with Fast Localized Spectral
                  Filtering},
  booktitle    = {Advances in Neural Information Processing Systems},
  pages        = {3837--3845},
  year         = {2016},
}

@inproceedings{2015Adam,
  author    = {Diederik P. Kingma and
               Jimmy Ba},
  title     = {Adam: {A} Method for Stochastic Optimization},
  booktitle = {International Conference on Learning Representations},
  year      = {2015},
  pages     = {1-15}
}

@article{2008T-SNE,
  author  = {Laurens van der Maaten and Geoffrey Hinton},
  title   = {Visualizing Data using t-SNE},
  journal = {J Mach Learn Res},
  year    = {2008},
  volume  = {9},
  number  = {86},
  pages   = {2579--2605},
}

@article{JGY1,
  title={Spatio-temporal graph neural networks for predictive learning in urban computing: A survey},
  author={Jin, Guangyin and Liang, Yuxuan and Fang, Yuchen and Shao, Zezhi and Huang, Jincai and Zhang, Junbo and Zheng, Yu},
  journal={IEEE Trans Knowl Data Eng},
  volume={36},
  number={10},
  pages={5388--5408},
  year={2023},
}

@article{JGY2,
  title={Automated dilated spatio-temporal synchronous graph modeling for traffic prediction},
  author={Jin, Guangyin and Li, Fuxian and Zhang, Jinlei and Wang, Mudan and Huang, Jincai},
  journal={IEEE Trans Intell Transp Syst},
  volume={24},
  number={8},
  pages={8820--8830},
  year={2022},
}

\newpage

\end{document}